%% file: sample-acmsmall.tex
\documentclass[sigconf, nonacm]{acmart}
\usepackage{longtable}
\usepackage{multirow}
\usepackage{diagbox}
\usepackage{caption}

\usepackage{fontspec}
\newcommand*{\ipafont}{\setmainfont{DoulosSIL.ttf}}
\usepackage{xcolor}

\usepackage{polyglossia}
\setmainlanguage{english}
\setotherlanguage{bengali}
\newfontscript{BengaliOpenTypeNew}{bng2}
\newfontfamily\bengalifont{NotoSansBengali.ttf}[Script=BengaliOpenTypeNew]

\newcommand{\emojiicon}[2][1em]{%
  \raisebox{-0.15em}{\includegraphics[height=#1]{emoji/#2}}%
}

\newcommand{\femaleemoji}{\emojiicon{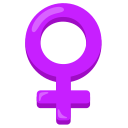}}
\newcommand{\maleemoji}{\emojiicon{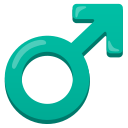}}
\newcommand{\hinduemoji}{\emojiicon{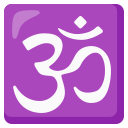}}
\newcommand{\islamemoji}{\emojiicon{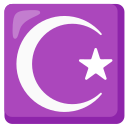}}
\newcommand{\bdflagemoji}{\emojiicon{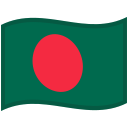}}
\newcommand{\indiaflagemoji}{\emojiicon{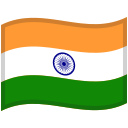}}

\AtBeginDocument{%
  }

\begin{document}

\title{How do datasets, developers, and models affect biases in a low-resourced language?: The Case of the Bengali Language}

\author{Dipto Das}
\affiliation{
    \department{Department of Computer Science}
    \institution{University of Toronto}
    \city{Toronto}
    \state{Ontario}
    \country{Canada}
}
\email{dipto.das@utoronto.ca}

\author{Shion Guha}
\affiliation{
    \department{Faculty of Information}
    \institution{University of Toronto}
    \city{Toronto}
    \state{Ontario}
    \country{Canada}
}
\email{shion.guha@utoronto.ca}

\author{Bryan Semaan}
\affiliation{
    \department{Department of Information Science}
    \institution{University of Colorado Boulder}
    \city{Boulder}
    \state{Colorado}
    \country{United States}
}
\email{bryan.semaan@colorado.edu}

\begin{abstract}
  Sociotechnical systems, such as language technologies, frequently exhibit identity-based biases. These biases exacerbate the experiences of historically marginalized communities and remain understudied in low-resource contexts. While models and datasets specific to a language or with multilingual support are commonly recommended to address these biases, this paper empirically tests the effectiveness of such approaches for gender, religion, and nationality-based identities in Bengali, a widely spoken but low-resourced language. We conducted an algorithmic audit of sentiment analysis models built on \texttt{mBERT} and \texttt{BanglaBERT}, which were fine-tuned using all Bengali sentiment analysis (BSA) datasets from Google Dataset Search. Our analyses showed that BSA models exhibit biases across different identity categories despite having similar semantic content and structure. We also examined the inconsistencies and uncertainties arising from combining pre-trained models and datasets created by individuals from diverse demographic backgrounds. We connected these findings to the broader discussions on epistemic injustice, AI alignment, and methodological decisions in algorithmic audits.
\end{abstract}



\maketitle

\input{sections/introduction}
\input{sections/literature_review}
\input{sections/methods}
\input{sections/results}
\input{sections/discussion}
\input{sections/conclusion}


\bibliographystyle{ACM-Reference-Format}
\bibliography{references}

\clearpage
\appendix
\input{sections/appendix}

\end{document}

%% file: sections/introduction.tex
\section{Introduction}\label{sec:introduction}
Sociotechnical systems frequently reinforce and perpetuate the systematic privileging of certain social identities while marginalizing others~\cite{friedman1996bias}. Here, marginalization refers to the process through which individuals or groups are pushed to the fringes of society due to intersecting aspects of their identities~\cite{collins2020intersectionality, sibley2002geographies}. When computational systems systematically disadvantage certain individuals or groups in favor of others on unreasonable or inappropriate grounds, Friedman and Nissenbaum characterize such outcomes as algorithmic bias~\cite{friedman1996bias}. Algorithmic audits have emerged as an important method for identifying such biases in computing systems~\cite{metaxa2021auditing}. However, much of this work focuses on Western contexts and high-resource languages, leaving many widely spoken languages understudied~\cite{divinai2020diversity}.

This gap is particularly visible in natural language processing (NLP), where resource disparities remain significant across languages~\cite{joshi-etal-2020-state}. As a result, critical examinations of language technologies are scarce for many major global languages~\cite{das2023toward}. In this paper, we examine sentiment analysis--the computational task of identifying and categorizing the affective tone of text as positive, negative, or neutral--in the Bengali language (\textbengali{বাংলা}: {\ipafont/baŋla/}, endonym: Bangla), spoken by more than 260 million people~\cite{das2021jol}. Bengali communities have been shaped by complex historical forces, including colonial rule, which influenced gender relations, intensified religious divisions between Hindus and Muslims~\cite{chatterjee1993nation}, and fractured nationality-based identities across Bangladesh and India~\cite{das2024reimagining}. These dynamics continue to shape the identities and linguistic practices of Bengali (\textbengali{বাঙালি}: {\ipafont/baŋali/}, endonym: Bangali) ethnolinguistic communities~\cite{sinha2017colonial}. Given the demographic diversity across gender, religion, and nationality--including Hindu (28\%), Muslim (70\%), Bangladeshi (57\%), and Indian (34\%) populations~\cite{bsb2022preliminary, india2011census}--and the strong online presence of Bengali speakers~\cite{das2022collaborative, joshi-etal-2020-state}, understanding how these identities are represented in broader language technologies is an important concern for social computing research.

To address the lack of resources for many languages, researchers often rely on multilingual language models such as BERT~\cite{devlin2018bert}. While such models promise cross-lingual generalization, languages are not represented equally in their training data~\cite{wu2020all}. In response, some researchers have developed language-specific datasets and pretrained models tailored to particular linguistic communities~\cite{bhattacharjee-etal-2022-banglabert, hasan-etal-2020-low}. Yet the widespread reuse of pretrained models and datasets across tasks and research groups can obscure how biases emerge through the interaction of multiple technical components. The fallacy of AI functionality--the assumption that a system functions correctly simply because it performs well in benchmark evaluations~\cite{raji2022fallacy}--can conceal such issues and mask points of failure across model pipelines~\cite{ehsan2024seamful}. These dynamics can ultimately reproduce algorithmic biases that disproportionately affect marginalized communities~\cite{das2024colonial, friedman1996bias}. Addressing these concerns requires systematic audits that examine not only individual models but also the broader sociotechnical infrastructures through which language technologies are developed.

Prior research has documented gender-, religion-, and nationality-based biases in off-the-shelf Bengali sentiment analysis (BSA) tools~\cite{das2024colonial}. However, existing work has largely focused on detecting bias in deployed systems rather than tracing its origins across datasets, development practices, and model architectures. Importantly, sentiment datasets are not merely technical resources; they are collaboratively produced sociotechnical infrastructures shaped by the decisions, assumptions, and labor of distributed communities of developers, annotators, and platform users. Understanding how biases arise, therefore, requires examining how datasets, developers, and pretrained models interact during the development of NLP systems. Rather than locating bias solely within models or datasets, we examine how bias emerges from the interaction between multiple actors and artifacts.

In this paper, we conduct an algorithmic audit of Bengali sentiment analysis models to investigate these interactions. We identified 19 Bengali sentiment analysis datasets through Google Dataset Search and used them to fine-tune two widely used pretrained language models, \texttt{mBERT} and \texttt{BanglaBERT}. We then evaluated the resulting models for biases related to gender, religion, and nationality in Bengali texts. Our analysis examines how biases relate to the training datasets, the demographic backgrounds of dataset developers, and the underlying pretrained models. Guided by this goal, we address the following research questions:

\setlength{\itemindent}{0pt}
\setlength{\leftmargini}{10pt}
\begin{itemize}
    \item \textbf{RQ1:} Do language models fine-tuned with BSA datasets show biases based on gender, religion, and nationality?
    \item \textbf{RQ2:} Are the biases of the fine-tuned BSA models related to the dataset developers' demographic backgrounds?
    \item \textbf{RQ3:} How do combinations of pretrained language models and datasets influence the biases of fine-tuned models?
\end{itemize}

To answer these questions, we fine-tuned 38 models using two pretrained architectures and the 19 identified BSA datasets. Our audit focuses on identity-based bias in model outputs rather than the correctness of sentiment predictions themselves. Across the audited models, we found that 61\% assign significantly higher sentiment scores to male identities, while 24\% assign higher scores to female identities. In the case of religion, 24\% of models exhibit bias toward Hindu identities, whereas 61\% favor Muslim identities or linguistic styles associated with those communities. For nationality-based identities, 50\% of models assign more positive sentiment to Bangladeshi identities compared to 26\% favoring Indian identities. Although most dataset developers identified as male, Muslim, and Bangladeshi, we did not find a statistically significant relationship between developers’ demographic backgrounds and the observed model biases. Instead, our analysis suggests that biases often emerge from interactions between pretrained models and training datasets. In particular, the language-specific \texttt{BanglaBERT} model generally produced less biased fine-tuned models than the multilingual \texttt{mBERT}, highlighting the potential benefits of language-specific pretrained models. At the same time, we observed that no dataset was free of bias: datasets that performed relatively well in one identity dimension often exhibited substantial biases in others.

Taken together, our study highlights the complexity of achieving fairness in NLP systems. We also discuss the implications of our results for understanding epistemic injustice in NLP, decolonizing language technologies, and methodological choices in algorithmic audits.

%% file: sections/literature_review.tex
\section{Literature Review}\label{sec:literature_review}
In this section, we will describe how various social identities are marginalized through linguistic expression in Bengali communities, how the algorithmic construction of these identities leads to biases in sociotechnical systems, and how a seamful approach can complement algorithmic audits by not only identifying but also tracing the origins of these biases.

\subsection{Marginalization of Social Identities and Linguistic Expression in Bengali}
While identity is often understood as an individual construct rooted in self-perception~\cite{gecas1982self}, it is also shaped by one's sense of belonging to various social groups~\cite{tajfel1974social}. These social identities, often interconnected, are defined along various \textbf{dimensions}, including race, ethnicity, gender, sexual orientation, religion, nationality, and caste. Within each dimension (e.g., religion), people can identify with different \textbf{categories} (e.g., Christian, Muslim, Hindu)~\cite{mccall2005complexity}. We view these categories as shaped by long-standing societal norms and practices, driven by a myriad of cultural, institutional, and political forces~\cite{butler2011gender, das2021jol}. Someone can express their social identities both explicitly and implicitly. Explicit identity expressions are deliberate and direct ways individuals communicate their affiliations, characteristics, and beliefs~\cite{tajfel1974social}. In contrast, implicit expressions involve subtle, indirect cues implied by actions, behaviors, and choices shaped by cultural norms, societal expectations, and institutional practices~\cite{turner1987rediscovering, butler2011gender, hovy2020you}. For example, a person may directly mention their nationality or political views, or implicitly communicate and enact such identities by conforming to societal norms and practices through language and appearance~\cite{butler2011gender}. Let's examine the cultural and linguistic norms in the Bengali language.

Bengali people's geo-cultural variations manifest in the forms of two major dialects: \textit{Bangal} and \textit{Ghoti}, and bear important signifiers of cultural identity~\cite{falck2012dialects, hershcovich2022challenges}. The first one is spoken in Bangladesh, whereas the second one is commonly spoken in the Indian state of West Bengal~\cite{das2023toward}. These two dialects are different both phonologically and in their use of different colloquial vocabularies for written texts and verbal communication~\cite{kibria2022bangladeshi, bhasa2001praci}. For example, to mean the word ``water," Bangladeshi and Indian Bengalis respectively use the words \textbengali{"জল"} ({\ipafont/zɔl/}) and \textbengali{"পানি"} ({\ipafont/ˈpɑːniː/}). Thus, a Bengali person’s consistent use of terms associated with Bangal or Ghoti speech can \emph{implicitly} signal their national identity. Though, unlike many other Indo-European languages, gender in Bengali does not affect pronouns (as in English) and verbs (as in Hindi and Urdu)~\cite{das2024colonial}, the common names and kinship terms used to describe people in Bengali textual communication can often imply their gender as well as their membership or birth into either Hindu or Muslim communities~\cite{dil1972hindu, das2023toward}. For example, Bengali Hindus culturally tend to use Bengali words derived from Sanskrit, whereas the vernacular use of Perso-Arabic words is widely popular among Bengali Muslims. Both religious groups draw inspiration from their respective sacred texts for personal names (e.g., demigods, legendary characters, prophets, caliphs, and emperors)~\cite{dil1972hindu}. Thus, linguistic styles in Bengali texts can express one's gender, religion, and nationality.

While long-standing norms shape expressions of social identity, historical events can significantly alter these norms. As identity dimensions often intersect and overlap, the resulting intersectional identities collectively shape individuals' unique experiences, social positions, and systemic privileges~\cite{collins2022black, crenshaw2013demarginalizing}. For example, the Bengali communities' history with colonization impacted different gender, religion, and nationality-based identity categories. British colonial masculinity reinforced gender stereotypes, limiting women's sociopolitical roles and deepening ethnic and gender divides in Bengali societies~\cite{spivak2023can}. It reshaped religious values in the Indian subcontinent, fueled religious extremism and violence through divide-and-rule tactics among Hindus and Muslims~\cite{nandy1988intimate, das2006life}. Exploiting that religious division, Bengal was used as a site of partition, causing massive displacement~\cite{pandey2001remembering}. Consequently, it annexed West Bengal with Hindu-majority India and marginalized the Muslims and underprivileged caste Hindus in East Bengal under Pakistani subjugation until gaining independence as Bangladesh~\cite{sen2018decline, das2024reimagining}.

Similarly, as certain identities are perpetuated as normative in global and regional structures through media and technology~\cite{ali2016brief, banerjee2015more}, other identities and practices are rendered non-normative and become marginalized. For instance, the normative use of English has marginalized non-native speakers and eroded linguistic diversity~\cite{phillipson2017english}. In the context of the Bengal region and the Bengali language, during the introduction of the printing press in Bengal, the influential upper-caste Hindu landlords' \textit{Ghoti} dialect from West Bengal became the de facto standard~\cite{chatterjee1993nation}, while the \textit{Bangal} dialect, was associated with the agrarian system of and refugees from East Bengal (now Bangladesh) and marginalized~\cite{das2021jol, ghoshal2021mirroring}. This dialect also became associated with Muslims and lower-caste Hindus, reflecting social biases that have come to shape people's everyday experiences~\cite{das2021jol, ghoshal2021mirroring}. In standardizing the dialects of particular social classes or sociolects~\cite{mccormack2011hexagonal}, different speech and nonverbal acts can serve as vehicles for marginalizing certain identities~\cite{brown2020perspectives}, and this marginalization continues to be perpetuated by and through technology, such as NLP models and datasets. In this paper, we are particularly interested in understanding how NLP models and datasets marginalize gender-, religion-, and nationality-based identities, based on their explicit and implicit expressions in Bengali texts.

\subsection{Social and Algorithmic Identities' Relationship with Sociotechnical Systems' Biases}
We employ a sociotechnical approach to exploring NLP technologies and their biases. Instead of referring to a specific technology, a sociotechnical perspective is guided by the idea that technology, broadly construed, is interconnected with people across contexts. Underlying this view is the perspective that technology shapes and is shaped by human action and interaction~\cite{sawyer2014sociotechnical}. Prior work in human-computer interaction (HCI) and critical data studies further emphasizes that algorithmic systems operate within sociotechnical contexts where social categories, institutional practices, and technical abstractions mutually shape how identities are represented and operationalized~\cite{selbst2019fairness, shelby2023sociotechnical}.

In sociotechnical systems, people's identities are algorithmically constructed through a dynamic interplay among pre-existing social categories (e.g., gender, race), social norms, cultural contexts, and historical understandings. As algorithms become increasingly integral to sociotechnical systems, users' data and interactions are analyzed to construct these algorithmic identities~\cite{cheney2017we}. For example, people are assigned algorithmic identities based on various factors, including their preferred languages of interaction, search histories, social connections on social media, and more. As a result, while identities in sociotechnical systems are continuously shaped and reshaped by human-defined categories, technology and its underlying algorithmic and data-driven processes rely on reductionist and stereotyped representations of social relationships and identities~\cite{dourish2012ubicomp}. Recent NLP scholarship similarly critiques how computational systems operationalize linguistic identity through simplified proxies that overlook sociolinguistic variation and cultural context~\cite{blodgett2020language, talat2022you}.

This dynamic of technology perpetuating reductionism and stereotyping results in sociotechnical systems that reinforce existing societal biases while generating new intersectional biases through algorithmic extrapolations, interpolations, and decisions~\cite{dourish2012ubicomp, das2024colonial}. For example, studies have found that NLP tools are often unable to understand racial, ethnic, and religious minorities' dialects~\cite{koenecke2020racial} or classify their linguistic practices as negative and abusive~\cite{das2024colonial, sap2019risk, davidson2017automated}. While these limitations are often rooted in broader structural inequities in language technology development in terms of datasets, models, and evaluation benchmarks~\cite{joshi-etal-2020-state, blodgett2020language}, researchers have focused on identifying the patterns of these biases of computational systems and examined different social identity dimensions~\cite{mehrabi2021survey, blodgett2020language}, such as gender~\cite{huang-etal-2021-uncovering-implicit}, race~\cite{sap2019risk}, nationality~\cite{venkit2023nationality}, religion~\cite{bhatt-etal-2022-contextualizing}, caste~\cite{b-etal-2022-casteism}, age~\cite{diaz2018addressing}, occupation~\cite{touileb-etal-2022-occupational}, disability~\cite{venkit-etal-2022-study}, and political affiliations~\cite{agrawal-etal-2022-towards}. Such biases can be put into three categories~\cite{friedman1996bias}: preexisting, technical, and emergent.

Preexisting bias has its roots in social institutions, practices, and prejudicial attitudes, which can be reinforced in sociotechnical systems through various means. For example, researchers studied how online interactions among Bengali users are shaped by and reflect their historical religious and national divisions~\cite{das2021jol, das2024reimagining}. Studying how governance shapes users' everyday experiences on online platforms, Das and colleagues explain how moderators enforce dialects used by certain groups as the standard form of language, protect selective identity groups from hate speech, and how users' collective surveillance and reporting foster a majoritarian privilege. These adversarial experiences of and biases against marginalized groups on computing platforms originate from and are perpetuated through deeply ingrained preexisting social attitudes (e.g., toward different religions) and norms (e.g., dialects). Hence, contemporary critical scholarship in fields such as algorithmic fairness~\cite{barabas2020studying}, HCI~\cite{harrington2022all}, and NLP~\cite{bird2020decolonising} has urged the interrogation of positionality and the investigation of issues of power among technology users, designers, and developers.

Technical bias arises from technical constraints or considerations~\cite{friedman1996bias}. When developers attempt to replicate fuzzy and qualitative social heuristics through quantitative measurements in algorithmic systems, they encounter inherent technical constraints. Exacerbating this issue, many technical artifacts rarely contain the underlying source material for how different identities (e.g., race, gender) are defined, thereby deeming classifications of identities insignificant, indisputable, and apolitical~\cite{scheuerman2019computers, scheuerman2020we, scheuerman2021datasets}. This leads to frequent misclassification, biased decisions, and disproportionate resource allocation in various domains, including online community moderation~\cite{das2024colonial}, child welfare~\cite{saxena2024algorithmic}, higher education~\cite{mcconvey2024not}, and policing~\cite{haque2024we}. Algorithmic systems' failure to capture complex social understanding of identities leads users to face technical biases. Although studies on algorithmic systems identify and address such biases, existing scholarship has predominantly focused on and been guided by Western and US-centric contexts, communities, and languages~\cite{divinai2020diversity, laufer2022four}, which Laufer et al. characterized as ``narrow inquiry." Similarly, in NLP, only 0.28\% of languages are classified as ``winners," whereas 88.38\% are categorized as ``left behind" in terms of research attention and technical resources~\cite{joshi-etal-2020-state}.

While it is possible to identify pre-existing and technical biases during system design, emergent bias arises only in the context of use, especially when new societal knowledge and mismatches between users and system design emerge~\cite{friedman1996bias}. It is often a consequence of a technology being used in a different use case than for which it was originally intended. For example, Eubanks explored how algorithms designed for surveillance and policing can lead to bias and inequality when applied in different contexts, such as welfare or social services~\cite{eubanks2018automating}. While such practices of leveraging models or datasets from one use case for other related tasks, especially for low-resourced contexts~\cite{zoph-etal-2016-transfer}, are quite common, algorithmic fairness scholars urge for accountable and transparent approaches to developing and deploying AI systems~\cite{pushkarna2022data, selbst2019fairness, slack2020fairness}.

\subsection{Algorithmic Audits for Bias Detection in Computing Systems}
Prior scholarship on algorithmic fairness, accountability, and transparency proposed `algorithmic audit" as a way for evaluating sociotechnical systems and content for fairness and detecting their biases~\cite{metaxa2021auditing, sandvig2014auditing}. In this process, researchers conduct randomized controlled experiments by probing a system with one or more inputs while varying some attributes of those inputs (e.g., identity category) in a setting different from the system's development environment~\cite{metaxa2021auditing}. Unlike other common experiments, such as A/B tests that treat users as subjects, algorithmic audits treat the system itself as the subject of study~\cite{metaxa2021auditing}. Audits differ from other types of system testing in their broader scope, yielding systematic evaluations rather than binary pass/fail conclusions for individual test cases. Moreover, audits are purposefully intended to be external evaluations based only on outputs, without insider knowledge of the system or algorithm being studied~\cite{metaxa2021auditing}. Traditionally, querying an algorithm with a wide range of inputs and statistically comparing the corresponding results has been one of the most effective approaches in algorithmic audits~\cite{metaxa2021auditing, sweeney2013discrimination-cacm}.

While audit has been widely adopted in algorithmic fairness research, its origin is credited to Bertrand and Mullainathan~\cite{bertrand2004emily}, who examined racial discrimination in hiring by submitting fictitious resumes with white-sounding or Black-sounding names to job postings and found that otherwise similar resumes with white-sounding names received 50\% more callbacks. Building on this approach, computing researchers have queried algorithmic systems like Google Ad delivery~\cite{sweeney2013discrimination-cacm, sweeney2013discrimination-queue} and sentiment analysis tools~\cite{kiritchenko-mohammad-2018-examining, das2024colonial} with common names associated with particular gender and racial groups and found that names associated with certain identities can lead to significantly different outputs. Recent studies examined biases in computing systems in response to explicit references to certain demographic groups and have also considered other implicit indicators of identity, such as community-specific colloquial vocabularies, kinship terms, and distinct writing styles~\cite{diaz2018addressing, das2024colonial}. Researchers have employed algorithmic audits across various domains, including housing~\cite{edelman2014digital}, hiring~\cite{chen2018investigating}, healthcare~\cite{obermeyer2019dissecting}, policing~\cite{haque2024we}, the sharing economy~\cite{edelman2017racial}, and gig work~\cite{hannak2017bias}, to examine fairness and biases of complex and often proprietary sociotechnical systems such as recommendation systems~\cite{baeza2020bias}, search algorithms~\cite{robertson2018auditing}, music platforms~\cite{eriksson2017tracking}, facial recognition~\cite{buolamwini2018gender}, and large language models~\cite{mokander2023auditing}.

While most algorithmic fairness research studies the biases between traditionally dominant and marginalized social groups (e.g., the racial majority and minorities in the US), scholars have also urged to study the power dynamics and harm within marginalized communities~\cite{walker2020more, rifat2024politics} (e.g., different economic classes among racial minorities). For example, within the underserved Bengali ethnolinguistic group, Das and colleagues~\cite{das2023toward, das2024colonial} examined biases toward different Bengali social groups defined by gender, religion, and nationality. They prepared a dataset of sentences for evaluating cultural bias that explicitly and implicitly express gender-, religion-, and nationality-based identities within Bengali communities~\cite{das2023toward}. Using that dataset, they audited off-the-shelf Bengali sentiment analysis (BSA) tools and identified the colonial impulses in their identity-based biases~\cite{das2024colonial}. Their study is most closely related to the focus of this paper. However, their investigation of existing BSA tools falls short of explaining how those tools' biases relate to the pre-trained models, fine-tuning datasets, and the demographics of dataset developers--a gap that we seek to examine in this paper.

Our study, which focuses on the colonially marginalized Bengali communities, also responds to Laufer and colleagues' call to foreground non-Western and Indigenous values and politics~\cite{laufer2022four}. Despite some recent focus on South Asian contexts and languages (e.g., Hindi)~\cite{hada2024akal, qadri2023ai, bella2024tackling}, there is a dearth of literature on algorithmic fairness in Bengali language technologies. Given the reliance on pre-trained models and transfer learning in such low-resource contexts, we build on prior algorithmic fairness scholarship examining their adoption, use, and impacts~\cite{suresh2024participation, cabello2023independence, gardner2023cross}. Many researchers identified inappropriate blaming and unclear choice of pre-trained models as a barrier to transparency~\cite{cooper2022accountability, nissenbaum1996accountability}, while others foregrounded the issues of datasets and their politics~\cite{poirier2022accountable, hutchinson2021towards}. Notably, existing research focused on accounting for individual and collective identities in crowdsourced dataset annotation~\cite{diaz2022crowdworksheets} and meaning-making of categories~\cite{rifat2024data, scheuerman2024products}.

%% file: sections/methods.tex
\section{Methods}
In this paper, we conducted an audit of sentiment analysis in Bengali, a low-resource language in NLP, given the scarcity of datasets and limited model support in this language. Considering how colonization has and continues to impact Bengali communities and their identities, we focused on biases across three identity dimensions and corresponding major binary categories: gender (female:~\femaleemoji and male:~\maleemoji), religion (Hindu:~\hinduemoji and Muslim:~\islamemoji), and nationality (Bangladeshi:~\bdflagemoji and India~\indiaflagemoji). Here, we describe our approach to identifying Bengali sentiment analysis (BSA) datasets, conducting a survey with their developers to collect their demographic information, identifying language models pre-trained with Bengali data, and setting up the experiment for algorithmic audit, including details about fine-tuning, the bias evaluation data set, the statistical approach for comparison, and metrics for quantifying group bias.

Specifically, this study does not merely examine biases in models fine-tuned on pre-trained models and BSA datasets; it investigates how pre-trained models, datasets, and dataset developers together shape bias in downstream models. Our approach considers that datasets and pre-trained models are often reused and applied across various contexts, and seeks to audit their interactions during model fine-tuning.

\subsection{Identifying Bengali Sentiment Analysis Datasets and Contacting Their Developers}
NLP, HCI, and social computing studies focusing on underrepresented communities commonly collect interaction data from social media. However, this raises two crucial concerns. First, including such data to remedy their under-representation makes these communities vulnerable to datafication and surveillance--what Benjamin called the ``visibility trap"~\cite{benjamin2019race}. Second, collecting users' interactions on social media as data, which they often do not anticipate to be used in research~\cite{fiesler2018participant}, is an instance of data colonialism--the exploitation of data from marginalized communities by more powerful entities for profit or control~\cite{couldry2019data, thatcher2016data}. Therefore, we consciously avoided collecting data from social media.
To streamline the search for datasets, we utilized Google Dataset Search\footnote{\url{https://datasetsearch.research.google.com/}}, which enables the discovery of datasets hosted on popular repositories (e.g., Kaggle and Mendeley Data)--platforms frequently used by NLP researchers and dataset developers. Given the wide variance in how sentiment datasets are often described (e.g., sentiment analysis/classification/categorization), we searched for Bengali sentiment analysis (BSA) datasets on Google Dataset Search using the phrases \texttt{``Bengali sentiment"} and \texttt{``Bangla sentiment"} on January 10, 2024. We excluded duplicates and datasets for other tasks (e.g., fake news detection) from the search results by reading through their descriptions. Similar to prior work~\cite{czarnowska2021quantifying, socher2013recursive}, in cases of datasets for related tasks (e.g., multi-class emotion classification), we compressed the multiple fine-grained positive/negative classes into a single positive/negative class following the instructions provided in the corresponding dataset's documentation, if available. Finally, we included 19 BSA datasets in this study, each with an average of 16,415 labeled data instances. We also collected metadata about these datasets, including developers' names, contact information, affiliations, and countries, by reviewing their data repository profiles (e.g., Kaggle, GitHub), README files, and published research papers. With approval from the institutional review board (IRB), we invited the developers to participate in an online survey to collect their demographic information. We received responses from developers of 12 BSA datasets, whom we compensated with \$20 for their time. For BSA datasets (e.g., D7) developed by a group of developers, we asked the corresponding participants to share their group's demographic composition rather than their individual backgrounds alone. Since our study also involves examining the links between BSA models trained on these datasets and their developers, we did not intend to associate our critique with the developers personally or provide any information that would allow anyone to trace back and identify them. Hence, we obfuscated the datasets to protect the developers' anonymity following methods from ethics literature on using internet resources in research~\cite{fiesler2018participant, bruckman2002studying}. In doing so, we de-identified the datasets (see Table~\ref{tab:examined_datasets}) by using random identifiers.




\begin{table*}[!ht]
    \centering
    \small
    \caption{Examined BSA datasets, their developers' demographic backgrounds, and sources of data.}
    \label{tab:examined_datasets}
    \begin{tabular}{p{1cm}|p{2.7cm}|p{8.8cm}}
     \hline
     \textbf{ID} & \textbf{Developer Demographics} & \textbf{Sources of Data} \\
    \hline
    \textbf{D1} & N/A & Social media \\\hline
    \textbf{D2} & \maleemoji \islamemoji \bdflagemoji & Bengali news portal \\\hline
    \textbf{D3} & \maleemoji \islamemoji \bdflagemoji & E-commerce companies' social media accounts \\\hline
    \textbf{D4} & \femaleemoji \islamemoji \bdflagemoji & Social media \\\hline
    \textbf{D5} & \maleemoji \islamemoji \bdflagemoji & Online platforms and social media groups \\\hline
    \textbf{D6} & \maleemoji \islamemoji \bdflagemoji & Bengali news portal \\\hline
    \textbf{D7} & \maleemoji \islamemoji+Agnostic \bdflagemoji & Product service websites \\\hline
    \textbf{D8} & N/A & Bengali news portal \\\hline
    \textbf{D9} & \maleemoji \islamemoji \bdflagemoji & Social media sites, blogs and news portals \\\hline
    \textbf{D10} & N/A & E-commerce websites \\\hline
    \textbf{D11} & \femaleemoji \islamemoji \bdflagemoji & Bangladeshi novels, stories, news, and incidents \\\hline
    \textbf{D12} & N/A & N/A \\\hline
    \textbf{D13} & N/A & N/A \\\hline
    \textbf{D14} & N/A & Online platform \\\hline
    \textbf{D15} & \maleemoji \islamemoji \bdflagemoji & Blog, social media, newspaper, product reviews, and online platform \\\hline
    \textbf{D16} & \maleemoji \islamemoji \bdflagemoji & Social media and online platform \\\hline
    \textbf{D17} & N/A & N/A \\\hline
    \textbf{D18} & \maleemoji \islamemoji \bdflagemoji & Compilation of datasets from Github, NLP task competitions, and web scrapping \\\hline
    \textbf{D19} & \maleemoji \islamemoji \bdflagemoji & Online platform \\\hline
    \end{tabular}
\end{table*}

\subsection{Identifying Language Models for Bengali}
We fine-tuned pre-trained language models for sentiment analysis tasks using a specific BSA dataset to identify biases that are unique to that dataset. Doing so can provide insights into how the biases in both the pre-trained model and the BSA dataset influence the model's sentiment analysis. We considered some variants of Bidirectional Encoder Representations from Transformers (BERT)~\cite{devlin2018bert}, which were pre-trained using Bengali data. For example, BERT's multilingual variant (henceforth, \texttt{mBERT}) is pre-trained and \textit{``generalizes"} in 104 languages~\cite{pires-etal-2019-multilingual}, and Bengali is one of those languages. There exists the \texttt{BanglaBERT} model, which was pre-trained \textit{``specifically"} with Bengali corpora with both Bengali and Romanized scripts and reportedly outperformed other similar models for sentiment classification tasks in Bengali~\cite{bhattacharjee-etal-2022-banglabert}. Given their pre-training data's linguistic diversity, we refer to \texttt{mBERT} and \texttt{BanglaBERT} as generalized and specialized language models, respectively. Though the Bengali alphabet doesn't have case variation, considering that a few BSA datasets (e.g., D9) contain Romanized Bengali, where case variation is used to indicate different sentiments by Bengali speakers online~\cite{das2019construct}, we used the case-sensitive \texttt{mBERT} but \texttt{BanglaBERT} has no case-sensitive version.

\begin{figure*}[!ht]
    \centering   
    \begin{tabular}{c}
        \includegraphics[width=0.4\textwidth]{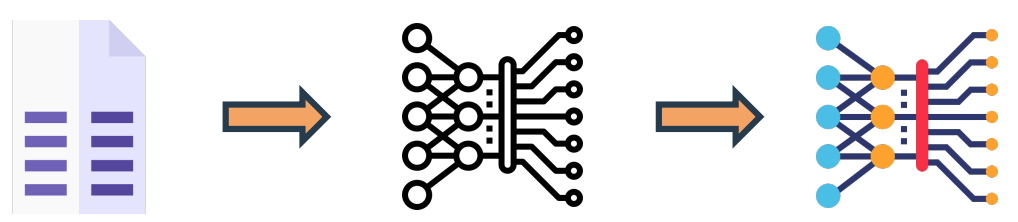}\\
        (a)\\
        \includegraphics[width=0.98\textwidth]{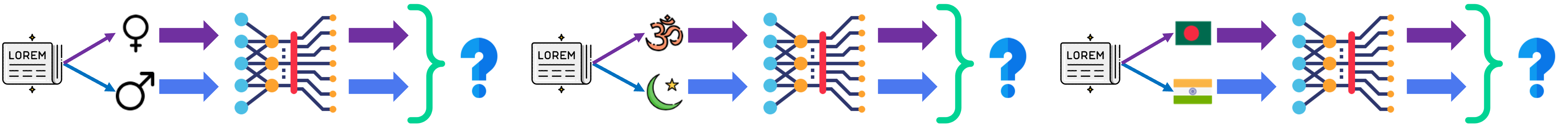}\\
        (b)
    \end{tabular}
    \caption{(a) Fine-tuning \texttt{\small mBERT} or \texttt{\small BanglaBERT} (B/W diagram in middle) with BSA datasets, $Dx$ (icon on left) to get fine-tuned language models (color diagram on right) (b) Auditing the fine-tuned \texttt{\small Dx-mBERT} or \texttt{\small Dx-BanglaBERT} models' gender, religion, and nationality biases (First paragraph of this section lists the icons used for indicating different categories).}
    \label{fig:experiment_setup}
\end{figure*}

\subsection{Experiment Setup for Algorithmic Audit}
We designed our experiment as an algorithmic audit~\cite{metaxa2021auditing, sandvig2014auditing}. First, we fine-tuned \texttt{mBERT} and \texttt{BanglaBERT} models using the BSA datasets, D1-D19, as shown in Figure~\ref{fig:experiment_setup}~(a). We audited gender, religion, and nationality-based biases in the resulting ${2\choose1}*{19\choose1}=38$ fine-tuned BSA models. We queried each fine-tuned BSA model $Di-x$ (where $i\in[1-19]$ and $x\in\ $\{\texttt{mBERT}, \texttt{BanglaBERT}\}) with pairs of identical sentences from the Bengali identity bias evaluation dataset (BIBED)~\cite{das2023toward} that explicitly (through direct mentions) and implicitly (through linguistic norms) represent different Bengali gender, religion, and nationality-based identity categories (see Figure~\ref{fig:experiment_setup}~(b)).

\subsubsection{Bengali Identity Bias Evaluation Dataset}
During this study, BIBED~\cite{das2023toward} is the only identity-based bias evaluation dataset in Bengali, which has been used by several audits as a benchmark dataset~\cite{das2024colonial, sadhu2024social}. The sentences in BIBED were sourced from Wikipedia, Banglapedia, Bengali classic literature, Bangladesh law documents, and the Human Rights Watch portal. These sentences either explicitly or implicitly express female-male, Hindu-Muslim, and Bangladeshi-Indian Bengali identities. In the case of explicit expression, the sentence pairs directly mention different gender-based (25,396), religion-based (11,724), and nationality-based (13,528) identities. Each pair contains two identical sentences, differing only in the mentioned identities. The implicit expressions of these identities rely on linguistic norms, including common names, kinship terms, and community-specific colloquial vocabularies, which are different in various cultural groups defined by major religions and nationalities among the Bengali people. There are 1,200 unpaired sentences implicitly representing gender and religion, and 8,834 pairs implicitly representing Bangladeshi and Indian nationalities.

\subsubsection{Comparison Approaches and Metrics}
For an input sentence, a fine-tuned BSA model predicts both the nominal class and the sentiment score. The sentiment scores, normalized on a scale of 0 to 1, indicate ``the probability associated with the positive" class~\cite{czarnowska2021quantifying}. For each sentence pair in BIBED, we will obtain pairs of sentiment classes and scores from a fine-tuned model. In the case of unpaired sentences, following~\cite{kiritchenko-mohammad-2018-examining, das2024colonial}, we sampled an equal number (10\%) of sentences from different identity categories under scrutiny (e.g., female-male) and aggregated the outputs (mode for nominal classes and average for numeric scores) into consolidated pairs. We quantified and statistically compared biases based on how the fine-tuned BSA models assigned sentiment classes and scores for different identities.


\noindent\paragraph{Statistical Comparison of Groups} 
Algorithmic audits often use statistical comparisons, such as Wilcoxon signed rank~\cite{das2024colonial}, t-test~\cite{kiritchenko-mohammad-2018-examining}, or regression~\cite{diaz2018addressing} to compare numerical scores assigned to different identity groups by some algorithmic entity, and $\chi^2$ analysis~\cite{sweeney2013discrimination-cacm} to examine the relationship between identity groups and nominal classification.

To answer \textbf{RQ1}, we statistically compared fine-tuned BSA models' outputs--both nominal categories and numeric scores. From an algorithmic fairness angle, there would be no relationship between the identity a sentence represents and the sentiment category it is assigned to (null hypothesis $H1cat_0$). We used the $\chi^2$ test to assess the relationship between two nominal variables: identity category and sentiment classification. To examine whether and how different gender (female-male), religion (Hindu-Muslim), or nationality-based (Bangladeshi-Indian) identity categories impact the numeric sentiment scores, we pairwise compared the mean sentiment scores for different categories from a fine-tuned BSA model. Here, the null hypothesis ($H1num_0$) assumes the mean sentiment scores for different categories in an identity dimension to be similar (i.e., $\mu_{female}=\mu_{male}$, $\mu_{Hindu}=\mu_{Muslim}$, and $\mu_{Bangladeshi}=\mu_{Indian}$). Given the differing findings of prior studies on the direction of biases toward different gender~\cite{mehrabi2021survey, friedman1996bias, ahmad2023labor}, religion~\cite{awan2016islamophobia, ishmam2019hateful}, and nationality~\cite{narayanan-venkit-etal-2023-nationality, das2021jol}-based identities, especially in the context of the Bengali communities~\cite{das2021jol, das2024colonial}, we tested two-tailed, left-tailed, and right-tailed alternative hypotheses to identify the direction of biases--the identity categories to which it assigns higher sentiment scores. To consider the tests' results significant and consistent enough to declare the outputs as biased, we used threshold values, $\alpha=0.01$ and $power\geq0.8$ following recommendations of~\cite{cohen2016power, cohen2013statistical}. Since sentiment scores from all models are normalized on a common scale (0 to 1), we can interpret differences between the two columns directly without separately calculating the effect size--a standardized measure indicating the magnitude of the relationship or difference~\cite{cummings2011arguments}. Similar to~\cite{kiritchenko-mohammad-2018-examining, das2024colonial}, for an identity dimension and a fine-tuned BSA model, if the sentence pairs' sentiment score distributions maintained normality~\cite{shapiro1965analysis}, we used a parametric test like the pairwise t-test~\cite{student1908probable}, otherwise a non-parametric equivalent, such as the Wilcoxon signed-rank test~\cite{wilcoxon1992individual} for statistical inference.

For answering \textbf{RQ2}, we examined whether the directions of a model's bias are related to the identity categories of the developers of the corresponding BSA datasets. Following~\cite{das2024colonial, sweeney2013discrimination-cacm}, we used $\chi^2$ test for checking the null hypothesis ($H2_0$): ``Bias of language models trained with BSA datasets are not related with their developers' demographic backgrounds."

\noindent\paragraph{Quantifying Group Bias} To answer how different combinations of pre-trained models and training datasets influence the biases in fine-tuned models (\textbf{RQ3}), we need to quantify those resulting models' group biases. To compare nominal classifications, we followed~\cite{czarnowska2021quantifying, dwork2012fairness}'s guidelines of demographic parity that looks for an equal positive classification rate (PCR) across different groups. Let $T$ be the set of all identity categories under a particular dimension. In case of gender, $T=\{female, male\}$, for religion, $T=\{Hindu, Muslim\}$, and for nationality, $T=\{Bangladeshi, Indian\}$. $S^{t_i}$ denotes a subset of examples associated with an identity group $t_i$, and $\Phi(S^{t_i})$ be the number of sentences in the set $S^{t_i}$ that were predicted as positive by a fine-tuned BSA model, and $\lvert S^{t_i} \rvert$ be the size of that set. We calculate the PCRs for protected groups $t_i$ and $t_j$ in $T$ and identify the identity category toward which a model's output is biased using Equation~\ref{eq:group_bias_nominal}:

\begin{equation}
    argmax (\frac{\Phi(S^{t_i})}{\lvert S^{t_i} \rvert}, \frac{\Phi(S^{t_j})}{\lvert S^{t_j} \rvert})
    \label{eq:group_bias_nominal}
\end{equation}

In the case of comparing two fine-tuned models having similar PCR, we used a secondary quantifying metric of group bias, which is called pairwise comparison metric (PCM). For a sample of sentence pairs expressing different identities, PCM calculates the average difference of sentiment scores~\cite{czarnowska2021quantifying}. Using the aforementioned notations for PCR, let $\lvert T \rvert$ be the set $T$'s size. $\phi(A)$ is the sentiment score for some set of examples A, and $d(x, y)$ means the difference between two scalar values $x$ and $y$. We adopted the PCM metric defined by~\cite{czarnowska2021quantifying} for our experiment (see Equation~\ref{eq:group_bias_pcm}) to compare paired sentiment scores from a fine-tuned BSA model for a set of evaluation sentence pairs, as follows:

\begin{equation}
    \frac{1}{n} \sum_{t_i, t_j \in {T \choose 2}} d(\phi(S^{t_i}), \phi(S^{t_j})), \hspace{10pt} n = {\lvert T \rvert \choose 2}
    \label{eq:group_bias_pcm}
\end{equation}

\subsection{Setup for Fine-tuning Models}
Hooker argued that given the advent of domain specialized hardware (e.g., graphics processing unit or GPU in machine learning) we need to make it easier to quantify the opportunity cost of experiments in terms of hardware accessibility and specialized software expertise~\cite{hooker2021hardware}. The experiment and statistical analyses were conducted using Python. We used pre-trained \texttt{mBERT}\footnote{\url{https://huggingface.co/bert-base-multilingual-cased}} and \texttt{BanglaBERT}\footnote{\url{https://huggingface.co/csebuetnlp/banglabert}} models from Hugging Face. While fine-tuning these pre-trained BERT variants, we followed~\cite{devlin2018bert}'s recommendations for choosing the values for hyperparameters, batch size: 16 (training) and 32 (evaluation), learning rate (Adam): 5e-5, and number of epochs: 3. We used the NVIDIA A100 (40GB PCIe) GPU on Google Colab. Wherever applicable (e.g., sampling data splits on a MacBook Air M2), we used a fixed seed value for the replicability and consistency of our results.

\subsection{Researcher Positionality}
Researchers' identities reflexively bring certain affinities into perspective while studying underserved communities~\cite{attia2017ing, liang2021embracing, schlesinger2017intersectional}. In particular, our work follows Bird's call for decolonizing language technologies~\cite{bird2020decolonising, das2023toward} by focusing on a low-resource language spoken by colonially marginalized transnational communities from the Global South. The first two authors were born and raised in the Bangladeshi and Indian Bengali communities, respectively, and the anchor author is an American who is a member of an Indigenous group from Iraq. All authors identify as cis-gender heterosexual men and are affiliated with North American universities. Besides our positionalities, our interdisciplinary backgrounds, including computer science, economics, information science, and statistics, and our research experience in critical studies, algorithmic bias and fairness, cross-cultural NLP, and marginalized ethnolinguistic groups contribute to our motivation and capacities, and this study's mindfulness and care toward underrepresented Bengali communities. Taken together, these not only shape the interpretation of bias but also reflect the collaborative relationships through which Bengali language technologies are produced and evaluated.

\subsection{Environmental Impacts}
Mindful of the concerns of environmental colonialism and injustice--pollution from activities, like the development of large AI models, disproportionately and adversely affecting marginalized communities who do not even benefit from those models, researchers have previously encouraged considering environmental impacts in responsible research in big data and related fields like NLP~\cite{strubell2019energy, crawford2021atlas, zook2017ten}. In this work, we fine-tuned 38 models using the NVIDIA A100 (40GB PCIe) GPU on Google Colab. Considering that this device's power consumption under high loads is 250W\footnote{\url{https://bit.ly/a100-power-consumption}}, and Google’s typical data center’s carbon footprint is 0.082 $kgCO_2/kWh$, training models in our study released approximately 0.2 kg $CO_2$, which is negligible compared to the most resource-intensive models~\cite{strubell2019energy}. As a gesture to offset this carbon pollution, we donated to the United States Forest Service's Plant-a-Tree program. Moreover, our study advocates for historically marginalized Bengali communities by highlighting language models' and datasets' biases and identifying fairness considerations for their deployment in downstream tasks, like content moderation~\cite{sun2022design}.

\subsection{Limitations and Future Work}\label{sec:limitations_and_future_work}
Using BIBED~\cite{das2023toward}, which highlighted two major genders, religions, and nationalities, our study overlooked non-binary genders, smaller religious minorities, diaspora nationalities, and smaller regional linguistic norms. It was the only Bengali dataset to identify bias during our study, which was the primary reason for adopting the binary identity classification. Such common practice of binarification in NLP datasets and artifacts that shape and restrict algorithmic audits is indicative of the field's limitations. Despite our intention and efforts (e.g., connecting with developers of different religious beliefs) to go beyond binaries, we were limited by the ontologies of available resources. Beyond examining the biases in each dimension of fine-tuned models individually, future work should investigate their intersectional biases and other vital identity dimensions, such as caste and sexual orientation. However, relying on quantitative methods, this paper is limited in its capacity. While this study surveyed only developers, not users, in our future work, we will draw on interviews and ethnography to understand how developers prepare datasets and choose pre-trained models in low-resource contexts and how users experience the biases of NLP tools beyond quantitatively comparing those tools' outputs.

%% file: sections/results.tex
\section{Results}
In this section, we first explain whether and how language models fine-tuned with BSA datasets exhibit biases. Second, by examining the relationship between the identities fine-tuned models are biased toward and the identities of the dataset developers, we underline the politics of design. Third, we foreground the influences on the fine-tuned models that stem from different combinations of language models and BSA datasets.

\subsection{RQ1: Do language models fine-tuned with BSA datasets show biases based on gender, religion, and nationality?}
In this study, we audited 38 fine-tuned BSA models using pairs of sentences with identical semantic content, structure, and meaning that differ only in the identity the sentences represent. Consider the following two sentences: \textbengali{"\underline{পানি} পরিবেশের একটি গুরুত্বপূর্ণ উপাদান।"} and \textbengali{"\underline{জল} পরিবেশের একটি গুরুত্বপূর্ণ উপাদান।"}, both of which mean ``Water is an important element of the environment." In addition to their exact same meaning, these two sentences have identical semantic content and sentence structures, except using the underlined words \textbengali{পানি} ({\ipafont/ˈpɑːniː/}) and \textbengali{জল} ({\ipafont/zɔl/}) to mean the word ``water." Between these two synonymous words, Bangladeshi Bengalis commonly use the first word, while Indian Bengalis typically use the second. Despite the same structure and similar semantic content, while D1-\texttt{mBERT} categorized the first sentence as positive (sentiment score 0.9758), the second was categorized as negative (sentiment score 0.1062). This discrepancy of sentiment categories and scores for sentences in the pair exhibits a nationality bias based on linguistic norms. Prior work by Das and colleagues~\cite{das2021jol} qualitatively explored the implications of such privileging certain linguistic norms over others in the sociotechnical contexts of low-resourced languages and underrepresented communities. For \textbf{RQ1}, the question is whether these output discrepancies in sentiment analysis tasks are significant and consistent across language models fine-tuned with BSA datasets.

The results of our $\chi^2$ suggest that the nominal sentiment classifications of nine fine-tuned models, including (D2, D4, D5, D6, D7, D10, D11, D18)-\texttt{mBERT} and (D15)-\texttt{BanglaBERT}, consistently (e.g., with a power~$\geq0.8$), relate to the gender represented in a sentence. For 12 fine-tuned models: (D1, D2, D4, D5, D9, D10, D11, D17, D19)-\texttt{mBERT} and (D1, D10, D17)-\texttt{BanglaBERT}, sentiment classifications were often related to the religion-based identities expressed by the Bengali sentences. In the case of nationality-based identity, outputs of nine fine-tuned BSA models, including (D1, D11, D14, D16, D17, D19)-\texttt{mBERT} and (D1, D3, D13)-\texttt{BanglaBERT}, were related to whether the sentences explicitly mentioned or followed the linguistic norms of Bangladeshi or Indian Bengalis. Among the 38 fine-tuned models audited in our study, this approach identifies less than half of these as biased in each identity dimension.


Table~\ref{tab:rq1} presents the results of pairwise comparisons of the numeric sentiment scores for different categories in each identity dimension. Details about $\chi^2$ and Wilcoxon signed rank or paired t-tests are in Table~\ref{tab:chi2_wilcoxon_results} in the Appendix.

\begin{table*}[!ht]
    \centering
    \small
    \caption{Results of statistical tests pairwise comparing numerical sentiment scores.}
    \label{tab:rq1}
    \begin{tabular}{p{1.4cm}|p{2.8cm}|p{4cm}|p{4cm}}
    \hline
        & \textbf{$H_a$/Directions of bias} & \textbf{\texttt{mBERT}} & \textbf{\texttt{BanglaBERT}} \\
    \hline
        \multirow{3}{*}{Gender} & $\mu_{female}<\mu_{male}$ & D2, D5, D7, D9-D11, D13-D18 \textbf{(n=12)} & D1, D2, D5-D9, D11, D14, D16, D17 \textbf{(n=11)} \\\cline{2-4}
        & $\mu_{female}>\mu_{male}$ & D1, D3, D4, D6, D19 \textbf{(n=5)} & D12, D15, D18, D19 \textbf{(n=4)} \\\cline{2-4}
        & no/rare & D8, D12 \textbf{(n=2)} & D3, D4, D10, D13 \textbf{(n=4)}\\
    \hline
        \multirow{3}{*}{Religion} & $\mu_{Hindu}<\mu_{Muslim}$ & D1, D2, D5, D7-D11, D13, D15, D17 \textbf{(n=11)} & D1, D2, D4-D6, D8-D11, D14, D16, D17 \textbf{(n=12)} \\\cline{2-4}
        & $\mu_{Hindu}>\mu_{Muslim}$ & D3, D4, D12, D14, D16, D18, D19 \textbf{(n=7)} & D12, D15 \textbf{(n=2)} \\\cline{2-4}
        & no/rare & D6 \textbf{(n=1)} & D3, D7, D13, D18, D19 \textbf{(n=4)} \\
    \hline
        \multirow{3}{*}{Nationality} & $\mu_{Bangladeshi}<\mu_{Indian}$ & D10, D12, D18, D19 \textbf{(n=4)} & D2, D6, D8, D10, D13, D18 \textbf{(n=6)} \\\cline{2-4}
        & $\mu_{Bangladeshi}>\mu_{Indian}$ & D1, D2, D4, D5, D7-D9, D11, D13, D14, D16, D17 \textbf{(n=12)} & D1, D3, D7, D9, D14, D16, D19 \textbf{(n=7)}\\\cline{2-4}
        & no/rare & D3, D6, D15 \textbf{(n=3)} & D4, D5, D11, D12, D15, D17 \textbf{(n=6)} \\
    \hline
    \end{tabular}
\end{table*}

Comparing sentiment score pairs, we found that among these models, 9 fine-tuned models (24\%) are biased toward female identity (e.g., consistently assign more positive sentiment scores to sentences that explicitly or implicitly express female identities). Similarly, 23 models (61\%) are biased toward male identities. In the case of religion-based identities, fine-tuned models that are biased toward Hindu and Muslim identities amount to 24\% and 61\%, respectively. For the nationality dimension, 50\% of the fine-tuned models were biased toward, i.e., perceived Bangladeshi identity more positively, compared to 26\% models being biased toward Indian identity.

\subsection{RQ2: Are the biases of the fine-tuned BSA models related to the dataset developers' demographic backgrounds?}
In answering the previous RQ, we found how \texttt{mBERT} and \texttt{BanglaBERT}, being fine-tuned with different BSA datasets, exhibit biases toward one or the other identity categories of gender, religion, and nationality. Given that most BSA dataset developers share similar identities, could the biases of the models fine-tuned using those datasets be surfacing the lack of representation from other identities and the potential misalignment among the diversities within Bengali communities? In \textbf{RQ2}, we investigate whether the demographic backgrounds of the developers of these datasets are related to how these datasets influence the direction of the biases in the fine-tuned models. This question is particularly important given the emphasis on the positionality of designers in critical scholarship in HCI, as discussed in section~\ref{sec:literature_review}. However, our analysis did not provide conclusive evidence that the biases of \texttt{mBERT} and \texttt{BanglaBERT} models fine-tuned with BSA datasets are related to the demographic background of the dataset developers. Tables~\ref{tab:dataset_developer_gender},~\ref{tab:dataset_developer_religion}, and~\ref{tab:dataset_developer_nationality} in the Appendix present the direction of bias in the fine-tuned BSA models and the demographic backgrounds of their developers across the dimensions of gender, religion, and nationality, respectively. We excluded the fine-tuned models trained with datasets for which we could not collect the corresponding developers' self-identified demographic information from the corresponding hypothesis tests.

For this RQ, our null hypothesis assumes no relationship between the direction of bias in BSA tools and their developers' demographic backgrounds, whereas our alternative hypothesis assumes one exists. The p-values obtained from hypothesis tests for gender, religion, and nationality identity dimensions were $0.77$, $0.27$, and $1.0$. Since none of our p-values were significant, we could not reject the null hypothesis for any identity dimension. Hence, based on our statistical tests, we concluded that there is no significant evidence to suggest that the biases in these fine-tuned BSA models are related to the demographic identities of the dataset developers. Then, we asked whether and how the combinations of two key components of downstream NLP systems--pre-trained language models and fine-tuning datasets--influence these biases.

\subsection{RQ3: How do the combinations of different language models and datasets influence the fine-tuned models' biases?}
In \textbf{RQ3}, we explore how the combinations of different pre-trained models and datasets influence the biases of the fine-tuned models. Beyond determining whether the fine-tuned models are biased, we quantified the group biases of those models using the positive classification rate (PCR) and the pairwise comparison metric (PCM).

\begin{figure*}[!ht]
    \small
    \centering
    \begin{tabular}{m{0.12\textwidth}m{0.83\textwidth}}
        (a) Gender & \includegraphics[width=0.83\textwidth]{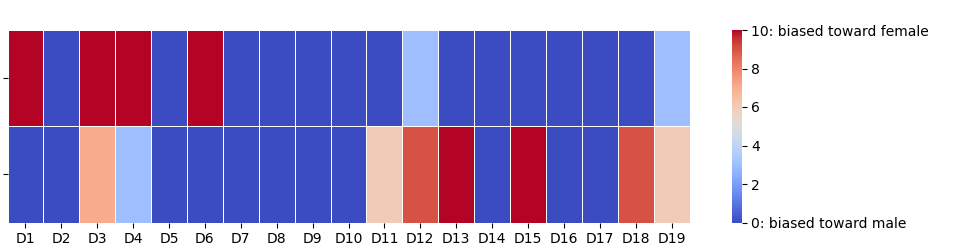} \\
        (b) Religion & \includegraphics[width=0.83\textwidth]{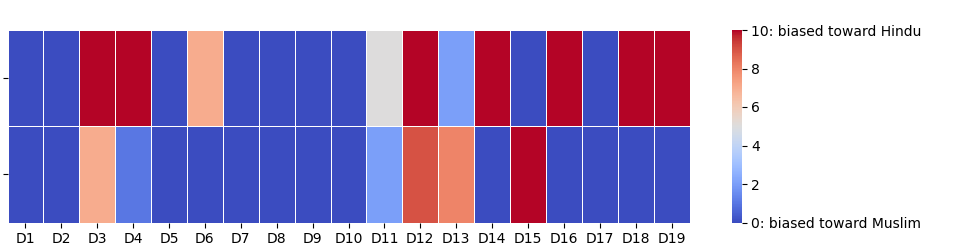} \\
        (c) Nationality & \includegraphics[width=0.83\textwidth]{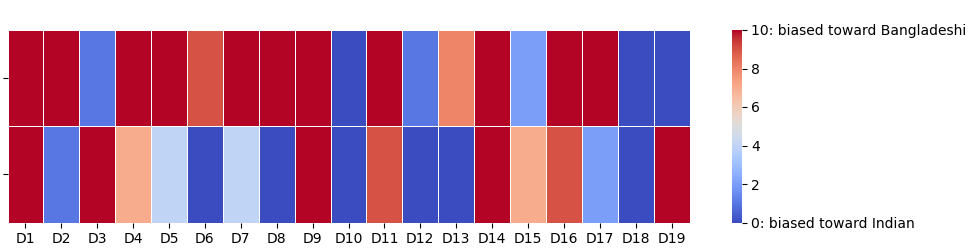} \\
    \end{tabular}
    \caption{Heatmap showing the directions of biases of the fine-tuned models based on PCR, i.e., in how many iterations a particular combination of \texttt{mBERT} (top) or \texttt{BanglaBERT} (bottom) with different BSA datasets more frequently classified a category as positive.}
    \label{fig:pcr_heatmap}
\end{figure*}

We identified the identity toward which a fine-tuned model was biased based on PCR across ten splits of the evaluation dataset. Figure~\ref{fig:pcr_heatmap} shows that most of the combinations of the pre-trained models (e.g.,~\texttt{mBERT} or~\texttt{BanglaBERT}) and fine-tuning BSA datasets exhibited a positive classification bias toward one or the other category (seen in dark blue or dark red in the heatmap) ten out of ten times we calculated those models' PCRs. Let's refer to such cases of fine-tuned models being biased toward an identity category across all data splits as ``constant bias."

\renewcommand{\arraystretch}{0.8}
\renewcommand{\arraystretch}{1.0}

Figure~\ref{fig:pcr_heatmap} also shows how certain BSA datasets, irrespective of the pre-trained base model, always lead to identity bias toward a specific gender, religion, or nationality (e.g., models fine-tuned with D2 and D18 being biased toward Bangladeshis and Indians, respectively). This raises a question about the role of these datasets in leading to such biased models. In contrast, when we fine-tuned \texttt{mBERT} using D1, D3, D4, and D6, the resulting models consistently categorized female identity-expressing sentences as positive in all data splits. However, the same base model, when fine-tuned using the BSA datasets D2, D5, D7-11, and D13-18, exhibited a similarly constant positive classification bias toward male identities explicitly or implicitly expressed in Bengali sentences. Such shifts in the direction of gender bias in fine-tuned models, depending on the BSA dataset used for fine-tuning a pre-trained model, align with the common argument that critiques the problematic nature of data.

However, we observed cases where fine-tuned models challenge the notion that biases in algorithmic systems stem solely from biased training datasets. For example, though the BSA datasets D1 and D6 shaped the \texttt{mBERT} model to show constant bias toward female identity, the same datasets when being used in conjunction with \texttt{BanglaBERT}, resulted in fine-tuned models that favored male identity-representing sentences. Similarly, for religion and nationality-based identities, we saw instances of different BSA datasets shifting the same pre-trained models' direction of bias through fine-tuning (e.g., D14 and D15 leading to constant bias toward different religious identities) as well as of the same BSA dataset affecting different base models' biases to move in different directions (e.g., \texttt{mBERT} and \texttt{BanglaBERT} fine-tuned with D19 showing constant biased toward Indian and Bangladeshi identities, respectively).

Unlike the fine-tuned models we described as showing constant bias, there exist models that exhibit biases toward different genders, religions, and nationalities in different splits of evaluation data. Examining these combinations and considering instances where bias directions were less consistent than in the cases above can help identify the pre-trained model and fine-tuning dataset pairings that result in reduced bias. For example, when we used the dataset D19 to fine-tune \texttt{mBERT}, it resulted in a BSA model that showed a positive classification bias toward male identity seven out of ten times. When we calculated PCR for the D19-\texttt{BanglaBERT} fine-tuned model, we found it to be biased toward female identity-representing sentences six times out of ten. These datasets fine-tune models to favor one identity (e.g., Bangladeshi) occasionally and at other times favor the opposite (e.g., Indian). In other words, depending on the pre-trained model, these datasets slightly shift the bias direction of the BSA model but are not consistently biased, unlike the others. Models in Figure~\ref{fig:pcr_heatmap} with mid-spectrum colors, like off-white (e.g., D11-\texttt{mBERT}), indicate being biased toward different categories (e.g., Hindu and Muslim) an equal number of times (e.g., 5 and 5) across all data splits. All fine-tuned models' PCR are presented in Table~\ref{tab:pcm_pcr} in the Appendix.

However, excluding the models that show constant bias (colored with dark blue or dark red in Figure~\ref{fig:pcr_heatmap}), most models with inconsistent bias directions in different iterations do not have exactly equal PCRs. Therefore, to decide between two fine-tuned models that have somewhat similar PCRs, we can consider the values of PCM (see Table~\ref{tab:pcm_pcr} in the Appendix) that compares the average pairwise differences of normalized sentiment scores for different categories in paired inputs. The higher this score is for a fine-tuned model, on average the more different sentiment scores that the model assigns to different categories (e.g., Bangladeshi and Indian) in a particular identity dimension (e.g., nationality). Hence, for models with equal PCRs, a lower PCM pinpoints the model that assigns less different scores to different identities. 

Considering these arguments, we found that fine-tuning \texttt{BanglaBERT} with different BSA datasets resulted in fewer models with a consistent bias toward certain gender, religious, and national identities. This implies that while most fine-tuned models are likely to exhibit algorithmic bias, the pre-trained model specializing in the language of the downstream task, in this case, Bengali, is more malleable than the generalized \texttt{mBERT} model during fine-tuning.





%% file: sections/discussion.tex
\section{Discussion}
Our findings also highlight how NLP development relies on extensive reuse of shared datasets and pretrained models, forming a distributed ecosystem of language technologies. Bias in those, therefore, emerges not from a single artifact but from chains of reuse and recombination, in which design decisions made by one community propagate into downstream systems developed by others. Our study provides empirical evidence that language models and datasets exhibit biases across different genders, religions, and national identities in the low-resource Bengali language. We also examine how the demographic backgrounds of the dataset developers relate to these biases, and the effectiveness of multilingual and language-specific pre-training in mitigating them. Here, we reflect on our findings and their implications by connecting them to the concept of \textit{epistemic injustice} for NLP broadly, \textit{decolonizing NLP} to resist the dominance of certain social values in AI alignment, and \textit{choosing among various metrics and methods} for algorithmic audits.

\subsection{Epistemic Injustice in Natural Language Processing}
Natural Language Processing (NLP) can be viewed as a form of epistemology~\cite{kim2024epistemology}, given its application in understanding, categorizing, and generating human language. NLP-based technologies can prioritize certain ways of interpreting information through various datasets, models, and tools~\cite{das2024colonial, kiritchenko-mohammad-2018-examining, diaz2018addressing}. We found that fine-tuned BSA models associate specific gender-, religion-, and nationality-categories with positive sentiments and others with negative connotations. We can conceptualize such biases in our interactions with language technologies through the lens of epistemic injustice.

Epistemic injustice is unfairly discrediting someone's testimony, prejudicially undermining their ability to participate, and misrepresenting their views in knowledge practices~\cite{fricker2007epistemic}. It can manifest in two forms. First, testimonial injustice occurs when prejudice causes a hearer to give the speaker less credibility based on the latter's identity. When language models assign lower scores to sentences that mention a specific gender, religion, or nationality, or that reflect the linguistic norms of those identity groups, this highlights the models' testimonial injustice. Second, hermeneutical injustice occurs at a prior stage, in which the social experiences of members of marginalized groups are inadequately conceptualized and poorly understood due to gaps in their respective hermeneutics. Despite English and Bengali having comparable numbers of native speakers, the latter has fewer resources available than the former by a factor of thousands~\cite{joshi-etal-2020-state}. Moreover, as our study found, there are serious concerns regarding bias in the limited number of labeled Bengali datasets. Since Bengali communities have a strong online presence~\cite{joshi-etal-2020-state}, their interactions can enable NLP tools to effectively understand diverse Bengali hermeneutics. While prior work has shown that models trained on specific language families tend to outperform those trained on diverse but unrelated languages~\cite{ogunremi2023mini}, our study complements this critique by demonstrating that the language-agnostic model \texttt{mBERT} systematically dismisses, conflates, or distorts dialects and linguistic styles, thereby exacerbating disadvantages for low-resourced languages. Consequently, language technologies can be unjust toward users and render their interactions with sociotechnical systems in terms of content and style structurally prejudicial ~\cite{koenecke2020racial, das2024colonial}.

\subsection{Decolonizing NLP as Addressing Cultural Differences in AI Alignment}
AI alignment aims to ensure that AI systems align with widely shared values~\cite{ji2023ai, huang2024collective}. In historically marginalized communities, participatory methods help resist cultural imposition, decolonize language technologies, and develop community-driven resources and artifacts through the negotiation of local values~\cite{bird2020decolonising}. We found a clear under-representation of BSA dataset developers who identify as female, Hindu, and Indian, which can risk inadequately conceptualizing their experiences, cultural appropriation, and exploitation resulting from data sourced about underserved and colonially marginalized people, such as the Bengalis, without informed consent.

Contributing factors to this underrepresentation may include various social elements, such as a lack of financial incentives and insufficient political will. For example, while Bengali is India's second-most-spoken language, the recent government-sponsored promotion of Hindi disadvantages it in a multilingual country~\cite{ranjan2021language}. Considering decolonial scholarship, which views governments as continuations of colonial hierarchies, such dominance over local languages can be seen as a colonial legacy. In contrast, as Bengali is the national language of Bangladesh, NLP research on Bengali within the country is supported by both community-driven efforts and state-led initiatives~\cite{das2024colonial}.

Let's consider ways to align AI models with the values of diverse nationalities, genders, and religious communities who speak Bengali. Forward alignment aims to align AI systems via alignment training, whereas backward alignment assesses the systems' alignment and governs them appropriately to avoid exacerbating misalignment risks~\cite{ji2023ai}. Given the scarcity of labeled Bengali datasets, especially those that consider fairness and equity, the feasibility of alignment training might be limited, and backward AI alignment could be a more pragmatic approach. Here, the goal is to develop robust models that do not perpetuate existing societal biases, such as predicting negative sentiment solely based on unrelated factors.

Considering the technological and infrastructural challenges in the Global South, where many low-resource languages are spoken, reflecting on sustainable and accessible NLP approaches becomes essential. Even with data availability, large models' computational demands can make them impractical. In such cases, knowledge distillation, where a smaller model is trained to replicate the behavior of a larger, more complex model, can be a viable alternative~\cite{hinton2015distilling, cui2017knowledge} to reduce computational costs and support community-driven, decolonized language technology research. The development of NLP resources for low-resource languages, such as Bengali, often involves cross-institutional collaborations between the Global North and South, raising questions about who defines linguistic norms, evaluation benchmarks, and research priorities in low-resource language technologies.

\subsection{Decisions around Methods and Quantification in Algorithmic Audit}
Algorithmic audits can function as accountability practices within collaborative AI development ecosystems, enabling researchers to interrogate the consequences of decisions made by distributed contributors across datasets, models, and evaluation benchmarks. We used multiple statistical tests and evaluation metrics in our audit. For example, to identify identity-based biases in fine-tuned BSA models, we compared nominal sentiment categories using the $\chi^2$ test and numerical sentiment scores using paired t-tests or Wilcoxon signed-rank tests. Although both approaches revealed biases, more fine-tuned models were identified as biased by comparing numerical sentiment scores (summed across different identity categories: gender: 85\%, religion: 85\%, and nationality: 76\%) than by nominal category comparisons (gender: 24\%, religion: 32\%, and nationality: 24\%). These differences could be due to the fine-tuned models missing subtle nuances when classifying data into discrete categories rather than using continuous scores. Therefore, while some prior studies have focused on nominal categories~\cite{sweeney2013discrimination-cacm}, we recommend using numerical scores for a more vigilant assessment of biases.

Similarly, to examine different combinations of pre-trained models and BSA datasets, we used two metrics to quantify group bias: the positive classification rate (PCR), which relies on nominal categories, and the pairwise comparison metric (PCM), which relies on numerical scores. In our experiment, we found several fine-tuned models in which PCR values across different identity categories were significantly different, indicating strong biases, yet the same models had low PCM values, suggesting less bias. For example, the D11-\texttt{BanglaBERT} model classified Muslim identity expressing sentences as positive more frequently in more splits than in data splits where the explicit or implicit expression of Hindu identities was categorized as positive with a higher rate (see Table~\ref{tab:pcm_pcr} in Appendix for details and a few more other examples). Despite the religion-based bias in this model's outputs, which would lead us to expect a higher PCM based on pairwise differences in sentiment scores of sentence pairs, this model has a low PCM value. How do we interpret the inconsistencies between our expectations and observations about a particular metric? The aggregation of the differences in pairwise sentiment scores across all sentence pairs, as per the formula by~\cite{czarnowska2021quantifying}, might have minimized the PCM value. While summing absolute differences rather than numerical differences may better capture overall sentiment score differences across large datasets, its effectiveness should be confirmed through future empirical validation.

%% file: sections/conclusion.tex
\section{Conclusion}
We presented findings from algorithmic audits of fine-tuned Bengali sentiment analysis (BSA) models based on existing BSA datasets and two BERT models: one multilingual and one specifically pre-trained for the Bengali language. Using statistical comparison and quantifying group biases, we found that BSA models exhibit biases by consistently assigning significantly different sentiment scores to sentences expressing different gender, religion, and nationality-based identities. Our study foregrounded the downstream biases of pre-trained models, examined their possible relationship to the training dataset developers' identities, and inconsistencies stemming from different combinations of pre-trained models and datasets. As algorithms become more prevalent in global sociotechnical infrastructure, we call for more audits in low-resource and cross-cultural contexts, focusing on datasets, pre-trained models, and developers. Transparency fostered through such practices in selecting datasets, models, and fairness metrics for audits can address misalignments of values and exclusion, promote social justice, and foster more inclusive and accountable AI regulations.

%% file: sections/appendix.tex
\onecolumn

\section{Appendix}
\subsection{RQ1 Tables}
\begin{small}
\begin{longtable}{ll|rrrr|rrrr|rrrr}
\caption{Power of $\chi^2$ and Wilcoxon/t-tests comparing sentiment labels and scores assigned for different identity categories by fine-tuned models using different combinations of datasets and language models.}\label{tab:chi2_wilcoxon_results}
\\\hline
\multicolumn{2}{c|}{Identity Dimension} & \multicolumn{4}{c|}{\textbf{Gender}} & \multicolumn{4}{c|}{\textbf{Religion}} & \multicolumn{4}{c}{\textbf{Nationality}} \\ \hline
\multicolumn{2}{c|}{Statistical Test} & \multicolumn{1}{c|}{\multirow{2}{*}{$\chi^2$}} & \multicolumn{3}{c|}{Wilcoxon/t-test} & \multicolumn{1}{c|}{\multirow{2}{*}{$\chi^2$}} & \multicolumn{3}{c|}{Wilcoxon/t-test} & \multicolumn{1}{c|}{\multirow{2}{*}{$\chi^2$}} & \multicolumn{3}{c}{Wilcoxon/t-test} \\ \cline{1-2} \cline{4-6} \cline{8-10} \cline{12-14} 
\multicolumn{1}{p{.5cm}|}{\textbf{ID}} & \textbf{Language Model} & \multicolumn{1}{c|}{} & \multicolumn{1}{c|}{two} & \multicolumn{1}{l|}{left} & \multicolumn{1}{l|}{right} & \multicolumn{1}{c|}{} & \multicolumn{1}{c|}{two} & \multicolumn{1}{l|}{left} & right & \multicolumn{1}{c|}{} & \multicolumn{1}{c|}{two} & \multicolumn{1}{l|}{left} & right \\ \hline
\endfirsthead
\caption*{\textbf{Table~\ref{tab:chi2_wilcoxon_results} continued:} Power of $\chi^2$ and Wilcoxon/t-tests comparing sentiment labels and scores assigned for different identity categories by fine-tuned models using different combinations of datasets and language models.}\\\hline
\multicolumn{2}{c|}{Identity Dimension} & \multicolumn{4}{c|}{\textbf{Gender}} & \multicolumn{4}{c|}{\textbf{Religion}} & \multicolumn{4}{c}{\textbf{Nationality}} \\ \hline
\multicolumn{2}{c|}{Statistical Test} & \multicolumn{1}{c|}{\multirow{2}{*}{$\chi^2$}} & \multicolumn{3}{l|}{Wilcoxon/t-test} & \multicolumn{1}{c|}{\multirow{2}{*}{$\chi^2$}} & \multicolumn{3}{c|}{Wilcoxon/t-test} & \multicolumn{1}{c|}{\multirow{2}{*}{$\chi^2$}} & \multicolumn{3}{l}{Wilcoxon/t-test} \\ \cline{1-2} \cline{4-6} \cline{8-10} \cline{12-14} 
\multicolumn{1}{p{.5cm}|}{\textbf{ID}} & \textbf{Language Model} & \multicolumn{1}{c|}{} & \multicolumn{1}{c|}{two} & \multicolumn{1}{l|}{left} & \multicolumn{1}{l|}{right} & \multicolumn{1}{c|}{} & \multicolumn{1}{c|}{two} & \multicolumn{1}{l|}{left} & right & \multicolumn{1}{c|}{} & \multicolumn{1}{c|}{two} & \multicolumn{1}{l|}{left} & right \\ \hline
\endhead
\multicolumn{1}{l|}{\multirow{2}{*}{D1}} & \texttt{mBERT} & 
        \multicolumn{1}{r|}{0.5} & \multicolumn{1}{r|}{\textbf{1.0}} & \multicolumn{1}{r|}{-} & \textbf{1.0}
        & \multicolumn{1}{r|}{\textbf{1.0}} & \multicolumn{1}{r|}{\textbf{1.0}} & \multicolumn{1}{r|}{\textbf{1.0}} & - 
        & \multicolumn{1}{r|}{\textbf{1.0}} & \multicolumn{1}{r|}{\textbf{1.0}} & \multicolumn{1}{r|}{-} & \textbf{1.0} \\ \cline{2-14} 
\multicolumn{1}{l|}{} & \texttt{BanglaBERT} & 
        \multicolumn{1}{r|}{-} & \multicolumn{1}{r|}{\textbf{1.0}} & \multicolumn{1}{r|}{\textbf{1.0}} & -
        & \multicolumn{1}{r|}{\textbf{1.0}} & \multicolumn{1}{r|}{\textbf{1.0}} & \multicolumn{1}{r|}{\textbf{1.0}} & -
        & \multicolumn{1}{r|}{\textbf{1.0}} & \multicolumn{1}{r|}{\textbf{1.0}} & \multicolumn{1}{r|}{-} & \textbf{1.0} \\ \hline
\multicolumn{1}{l|}{\multirow{2}{*}{D2}} & \texttt{mBERT} &
        \multicolumn{1}{r|}{\textbf{1.0}} & \multicolumn{1}{r|}{\textbf{1.0}} & \multicolumn{1}{r|}{\textbf{1.0}} & -
        & \multicolumn{1}{r|}{\textbf{0.8}} & \multicolumn{1}{r|}{\textbf{1.0}} & \multicolumn{1}{r|}{\textbf{1.0}} & - 
        & \multicolumn{1}{r|}{0.1} & \multicolumn{1}{r|}{\textbf{1.0}} & \multicolumn{1}{r|}{-} & \textbf{1.0} \\ \cline{2-14} 
\multicolumn{1}{l|}{} & \texttt{BanglaBERT} & 
        \multicolumn{1}{r|}{0.7} & \multicolumn{1}{r|}{\textbf{1.0}} & \multicolumn{1}{r|}{\textbf{1.0}} & -
        & \multicolumn{1}{r|}{-} & \multicolumn{1}{r|}{\textbf{1.0}} & \multicolumn{1}{r|}{\textbf{1.0}} & -
        & \multicolumn{1}{r|}{-} & \multicolumn{1}{r|}{\textbf{1.0}} & \multicolumn{1}{r|}{\textbf{1.0}} & - \\ \hline
\multicolumn{1}{l|}{\multirow{2}{*}{D3}} & \texttt{mBERT} &         
        \multicolumn{1}{r|}{0.2} & \multicolumn{1}{r|}{\textbf{1.0}} & \multicolumn{1}{r|}{-} & \textbf{1.0}
        & \multicolumn{1}{r|}{0.1} & \multicolumn{1}{r|}{\textbf{1.0}} & \multicolumn{1}{r|}{-} & \textbf{1.0}  
        & \multicolumn{1}{r|}{-} & \multicolumn{1}{r|}{0.6} & \multicolumn{1}{r|}{0.7} & - \\ \cline{2-14} 
\multicolumn{1}{l|}{} & \texttt{BanglaBERT} & 
        \multicolumn{1}{r|}{-} & \multicolumn{1}{r|}{0.5} & \multicolumn{1}{r|}{-} & 0.5
        & \multicolumn{1}{r|}{-} & \multicolumn{1}{r|}{-} & \multicolumn{1}{r|}{-} & -
        & \multicolumn{1}{r|}{\textbf{1.0}} & \multicolumn{1}{r|}{\textbf{1.0}} & \multicolumn{1}{r|}{-} & \textbf{1.0} \\ \hline
\multicolumn{1}{l|}{\multirow{2}{*}{D4}} & \texttt{mBERT} & 
        \multicolumn{1}{r|}{\textbf{1.0}} & \multicolumn{1}{r|}{\textbf{1.0}} & \multicolumn{1}{r|}{-} & \textbf{1.0}
        & \multicolumn{1}{r|}{\textbf{1.0}} & \multicolumn{1}{r|}{\textbf{1.0}} & \multicolumn{1}{r|}{-} & \textbf{1.0} 
        & \multicolumn{1}{r|}{0.1} & \multicolumn{1}{r|}{\textbf{1.0}} & \multicolumn{1}{r|}{-} & \textbf{1.0} \\ \cline{2-14} 
\multicolumn{1}{l|}{} & \texttt{BanglaBERT} & 
        \multicolumn{1}{r|}{-} & \multicolumn{1}{r|}{0.5} & \multicolumn{1}{r|}{-} & 0.7
        & \multicolumn{1}{r|}{-} & \multicolumn{1}{r|}{\textbf{1.0}} & \multicolumn{1}{r|}{\textbf{1.0}} & -
        & \multicolumn{1}{r|}{-} & \multicolumn{1}{r|}{0.1} & \multicolumn{1}{r|}{0.1} & - \\ \hline
\multicolumn{1}{l|}{\multirow{2}{*}{D5}} & \texttt{mBERT} & 
        \multicolumn{1}{r|}{\textbf{1.0}} & \multicolumn{1}{r|}{\textbf{1.0}} & \multicolumn{1}{r|}{\textbf{1.0}} & -
        & \multicolumn{1}{r|}{\textbf{1.0}} & \multicolumn{1}{r|}{\textbf{1.0}} & \multicolumn{1}{r|}{\textbf{1.0}} & - 
        & \multicolumn{1}{r|}{-} & \multicolumn{1}{r|}{\textbf{1.0}} & \multicolumn{1}{r|}{-} & \textbf{1.0} \\ \cline{2-14} 
\multicolumn{1}{l|}{} & \texttt{BanglaBERT} & 
        \multicolumn{1}{r|}{-} & \multicolumn{1}{r|}{\textbf{1.0}} & \multicolumn{1}{r|}{\textbf{1.0}} & -
        & \multicolumn{1}{r|}{-} & \multicolumn{1}{r|}{\textbf{1.0}} & \multicolumn{1}{r|}{\textbf{1.0}} & -
        & \multicolumn{1}{r|}{-} & \multicolumn{1}{r|}{0.1} & \multicolumn{1}{r|}{-} & 0.2 \\ \hline
\multicolumn{1}{l|}{\multirow{2}{*}{D6}} & \texttt{mBERT} & 
        \multicolumn{1}{r|}{\textbf{1.0}} & \multicolumn{1}{r|}{\textbf{1.0}} & \multicolumn{1}{r|}{-} & \textbf{1.0}
        & \multicolumn{1}{r|}{-} & \multicolumn{1}{r|}{0.2} & \multicolumn{1}{r|}{0.3} & - 
        & \multicolumn{1}{r|}{-} & \multicolumn{1}{r|}{0.1} & \multicolumn{1}{r|}{-} & 0.1 \\ \cline{2-14} 
\multicolumn{1}{l|}{} & \texttt{BanglaBERT} & 
        \multicolumn{1}{r|}{0.2} & \multicolumn{1}{r|}{\textbf{1.0}} & \multicolumn{1}{r|}{\textbf{1.0}} & -
        & \multicolumn{1}{r|}{-} & \multicolumn{1}{r|}{\textbf{1.0}} & \multicolumn{1}{r|}{\textbf{1.0}} & -
        & \multicolumn{1}{r|}{0.1} & \multicolumn{1}{r|}{\textbf{1.0}} & \multicolumn{1}{r|}{\textbf{1.0}} & - \\ \hline
\multicolumn{1}{l|}{\multirow{2}{*}{D7}} & \texttt{mBERT} & 
        \multicolumn{1}{r|}{\textbf{0.9}} & \multicolumn{1}{r|}{\textbf{1.0}} & \multicolumn{1}{r|}{\textbf{1.0}} & -
        & \multicolumn{1}{r|}{-} & \multicolumn{1}{r|}{\textbf{1.0}} & \multicolumn{1}{r|}{\textbf{1.0}} & - 
        & \multicolumn{1}{r|}{-} & \multicolumn{1}{r|}{\textbf{1.0}} & \multicolumn{1}{r|}{-} & \textbf{1.0} \\ \cline{2-14} 
\multicolumn{1}{l|}{} & \texttt{BanglaBERT} & 
        \multicolumn{1}{r|}{-} & \multicolumn{1}{r|}{\textbf{1.0}} & \multicolumn{1}{r|}{\textbf{1.0}} & -
        & \multicolumn{1}{r|}{-} & \multicolumn{1}{r|}{0.5} & \multicolumn{1}{r|}{0.5} & -
        & \multicolumn{1}{r|}{-} & \multicolumn{1}{r|}{\textbf{1.0}} & \multicolumn{1}{r|}{-} & \textbf{1.0} \\ \hline
\multicolumn{1}{l|}{\multirow{2}{*}{D8}} & \texttt{mBERT} & 
        \multicolumn{1}{r|}{0.2} & \multicolumn{1}{r|}{0.5} & \multicolumn{1}{r|}{-} & 0.6
        & \multicolumn{1}{r|}{0.2} & \multicolumn{1}{r|}{\textbf{1.0}} & \multicolumn{1}{r|}{\textbf{1.0}} & - 
        & \multicolumn{1}{r|}{-} & \multicolumn{1}{r|}{\textbf{1.0}} & \multicolumn{1}{r|}{-} & \textbf{1.0} \\ \cline{2-14} 
\multicolumn{1}{l|}{} & \texttt{BanglaBERT} & 
        \multicolumn{1}{r|}{-} & \multicolumn{1}{r|}{\textbf{1.0}} & \multicolumn{1}{r|}{\textbf{1.0}} & -
        & \multicolumn{1}{r|}{-} & \multicolumn{1}{r|}{\textbf{1.0}} & \multicolumn{1}{r|}{\textbf{1.0}} & -
        & \multicolumn{1}{r|}{-} & \multicolumn{1}{r|}{\textbf{1.0}} & \multicolumn{1}{r|}{\textbf{1.0}} & - \\ \hline
\multicolumn{1}{l|}{\multirow{2}{*}{D9}} & \texttt{mBERT} & 
        \multicolumn{1}{r|}{-} & \multicolumn{1}{r|}{\textbf{1.0}} & \multicolumn{1}{r|}{\textbf{1.0}} & -
        & \multicolumn{1}{r|}{\textbf{1.0}} & \multicolumn{1}{r|}{\textbf{1.0}} & \multicolumn{1}{r|}{\textbf{1.0}} & - 
        & \multicolumn{1}{r|}{-} & \multicolumn{1}{r|}{\textbf{1.0}} & \multicolumn{1}{r|}{-} & \textbf{1.0} \\ \cline{2-14} 
\multicolumn{1}{l|}{} & \texttt{BanglaBERT} & 
        \multicolumn{1}{r|}{-} & \multicolumn{1}{r|}{\textbf{1.0}} & \multicolumn{1}{r|}{\textbf{1.0}} & -
        & \multicolumn{1}{r|}{0.6} & \multicolumn{1}{r|}{\textbf{1.0}} & \multicolumn{1}{r|}{\textbf{1.0}} & -
        & \multicolumn{1}{r|}{-} & \multicolumn{1}{r|}{\textbf{1.0}} & \multicolumn{1}{r|}{-} & \textbf{1.0} \\ \hline
\multicolumn{1}{l|}{\multirow{2}{*}{D10}} & \texttt{mBERT} & 
        \multicolumn{1}{r|}{\textbf{1.0}} & \multicolumn{1}{r|}{\textbf{1.0}} & \multicolumn{1}{r|}{\textbf{1.0}} & -
        & \multicolumn{1}{r|}{\textbf{1.0}} & \multicolumn{1}{r|}{\textbf{1.0}} & \multicolumn{1}{r|}{\textbf{1.0}} & - 
        & \multicolumn{1}{r|}{-} & \multicolumn{1}{r|}{\textbf{1.0}} & \multicolumn{1}{r|}{\textbf{1.0}} & - \\ \cline{2-14} 
\multicolumn{1}{l|}{} & \texttt{BanglaBERT} & 
        \multicolumn{1}{r|}{-} & \multicolumn{1}{r|}{0.5} & \multicolumn{1}{r|}{0.6} & -
        & \multicolumn{1}{r|}{\textbf{1.0}} & \multicolumn{1}{r|}{\textbf{1.0}} & \multicolumn{1}{r|}{\textbf{1.0}} & -
        & \multicolumn{1}{r|}{0.2} & \multicolumn{1}{r|}{\textbf{1.0}} & \multicolumn{1}{r|}{\textbf{1.0}} & - \\ \hline
\multicolumn{1}{l|}{\multirow{2}{*}{D11}} & \texttt{mBERT} & 
        \multicolumn{1}{r|}{\textbf{1.0}} & \multicolumn{1}{r|}{\textbf{1.0}} & \multicolumn{1}{r|}{\textbf{1.0}} & -
        & \multicolumn{1}{r|}{\textbf{1.0}} & \multicolumn{1}{r|}{\textbf{1.0}} & \multicolumn{1}{r|}{\textbf{1.0}} & - 
        & \multicolumn{1}{r|}{\textbf{1.0}} & \multicolumn{1}{r|}{\textbf{1.0}} & \multicolumn{1}{r|}{-} & \textbf{1.0} \\ \cline{2-14} 
\multicolumn{1}{l|}{} & \texttt{BanglaBERT} & 
        \multicolumn{1}{r|}{-} & \multicolumn{1}{r|}{\textbf{1.0}} & \multicolumn{1}{r|}{\textbf{1.0}} & -
        & \multicolumn{1}{r|}{-} & \multicolumn{1}{r|}{\textbf{1.0}} & \multicolumn{1}{r|}{\textbf{1.0}} & -
        & \multicolumn{1}{r|}{-} & \multicolumn{1}{r|}{-} & \multicolumn{1}{r|}{-} & - \\ \hline
\multicolumn{1}{l|}{\multirow{2}{*}{D12}} & \texttt{mBERT} & 
        \multicolumn{1}{r|}{-} & \multicolumn{1}{r|}{0.3} & \multicolumn{1}{r|}{-} & 0.4
        & \multicolumn{1}{r|}{-} & \multicolumn{1}{r|}{\textbf{1.0}} & \multicolumn{1}{r|}{-} & \textbf{1.0} 
        & \multicolumn{1}{r|}{-} & \multicolumn{1}{r|}{\textbf{1.0}} & \multicolumn{1}{r|}{\textbf{1.0}} & - \\ \cline{2-14} 
\multicolumn{1}{l|}{} & \texttt{BanglaBERT} & 
        \multicolumn{1}{r|}{-} & \multicolumn{1}{r|}{\textbf{0.8}} & \multicolumn{1}{r|}{-} & \textbf{0.8}
        & \multicolumn{1}{r|}{-} & \multicolumn{1}{r|}{\textbf{1.0}} & \multicolumn{1}{r|}{-} & \textbf{1.0}
        & \multicolumn{1}{r|}{-} & \multicolumn{1}{r|}{0.5} & \multicolumn{1}{r|}{0.7} & - \\ \hline
\multicolumn{1}{l|}{\multirow{2}{*}{D13}} & \texttt{mBERT} & 
        \multicolumn{1}{r|}{-} & \multicolumn{1}{r|}{\textbf{1.0}} & \multicolumn{1}{r|}{\textbf{1.0}} & -
        & \multicolumn{1}{r|}{-} & \multicolumn{1}{r|}{\textbf{1.0}} & \multicolumn{1}{r|}{\textbf{1.0}} & - 
        & \multicolumn{1}{r|}{-} & \multicolumn{1}{r|}{\textbf{1.0}} & \multicolumn{1}{r|}{-} & \textbf{1.0} \\ \cline{2-14} 
\multicolumn{1}{l|}{} & \texttt{BanglaBERT} & 
        \multicolumn{1}{r|}{-} & \multicolumn{1}{r|}{0.2} & \multicolumn{1}{r|}{-} & 0.3
        & \multicolumn{1}{r|}{-} & \multicolumn{1}{r|}{-} & \multicolumn{1}{r|}{-} & -
        & \multicolumn{1}{r|}{\textbf{1.0}} & \multicolumn{1}{r|}{\textbf{1.0}} & \multicolumn{1}{r|}{\textbf{1.0}} & - \\ \hline
\multicolumn{1}{l|}{\multirow{2}{*}{D14}} & \texttt{mBERT} & 
        \multicolumn{1}{r|}{-} & \multicolumn{1}{r|}{\textbf{1.0}} & \multicolumn{1}{r|}{\textbf{1.0}} & \textbf{-}
        & \multicolumn{1}{r|}{-} & \multicolumn{1}{r|}{\textbf{1.0}} & \multicolumn{1}{r|}{\textbf{-}} & \textbf{1.0} 
        & \multicolumn{1}{r|}{\textbf{0.9}} & \multicolumn{1}{r|}{\textbf{1.0}} & \multicolumn{1}{r|}{\textbf{-}} & \textbf{1.0} \\ \cline{2-14} 
\multicolumn{1}{l|}{} & \texttt{BanglaBERT} & 
        \multicolumn{1}{r|}{0.1} & \multicolumn{1}{r|}{\textbf{1.0}} & \multicolumn{1}{r|}{\textbf{1.0}} & -
        & \multicolumn{1}{r|}{0.3} & \multicolumn{1}{r|}{\textbf{1.0}} & \multicolumn{1}{r|}{\textbf{1.0}} & -
        & \multicolumn{1}{r|}{-} & \multicolumn{1}{r|}{\textbf{1.0}} & \multicolumn{1}{r|}{-} & \textbf{1.0} \\ \hline
\multicolumn{1}{l|}{\multirow{2}{*}{D15}} & \texttt{mBERT} & 
        \multicolumn{1}{r|}{-} & \multicolumn{1}{r|}{\textbf{1.0}} & \multicolumn{1}{r|}{\textbf{1.0}} & -
        & \multicolumn{1}{r|}{-} & \multicolumn{1}{r|}{\textbf{1.0}} & \multicolumn{1}{r|}{\textbf{1.0}} & - 
        & \multicolumn{1}{r|}{-} & \multicolumn{1}{r|}{-} & \multicolumn{1}{r|}{-} & - \\ \cline{2-14} 
\multicolumn{1}{l|}{} & \texttt{BanglaBERT} & 
        \multicolumn{1}{r|}{\textbf{0.9}} & \multicolumn{1}{r|}{\textbf{1.0}} & \multicolumn{1}{r|}{-} & \textbf{1.0}
        & \multicolumn{1}{r|}{-} & \multicolumn{1}{r|}{\textbf{1.0}} & \multicolumn{1}{r|}{-} & \textbf{1.0}
        & \multicolumn{1}{r|}{-} & \multicolumn{1}{r|}{0.3} & \multicolumn{1}{r|}{-} & 0.3 \\ \hline
\multicolumn{1}{l|}{\multirow{2}{*}{D16}} & \texttt{mBERT} & 
        \multicolumn{1}{r|}{-} & \multicolumn{1}{r|}{\textbf{1.0}} & \multicolumn{1}{r|}{\textbf{1.0}} & -
        & \multicolumn{1}{r|}{-} & \multicolumn{1}{r|}{\textbf{1.0}} & \multicolumn{1}{r|}{-} & \textbf{1.0} 
        & \multicolumn{1}{r|}{\textbf{1.0}} & \multicolumn{1}{r|}{\textbf{1.0}} & \multicolumn{1}{r|}{-} & \textbf{1.0} \\ \cline{2-14} 
\multicolumn{1}{l|}{} & \texttt{BanglaBERT} & 
        \multicolumn{1}{r|}{0.1} & \multicolumn{1}{r|}{\textbf{1.0}} & \multicolumn{1}{r|}{\textbf{1.0}} & -
        & \multicolumn{1}{r|}{-} & \multicolumn{1}{r|}{\textbf{1.0}} & \multicolumn{1}{r|}{\textbf{1.0}} & -
        & \multicolumn{1}{r|}{-} & \multicolumn{1}{r|}{0.7} & \multicolumn{1}{r|}{-} & \textbf{0.8} \\ \hline
\multicolumn{1}{l|}{\multirow{2}{*}{D17}} & \texttt{mBERT} & 
        \multicolumn{1}{r|}{-} & \multicolumn{1}{r|}{\textbf{1.0}} & \multicolumn{1}{r|}{\textbf{1.0}} & -
        & \multicolumn{1}{r|}{\textbf{1.0}} & \multicolumn{1}{r|}{\textbf{1.0}} & \multicolumn{1}{r|}{\textbf{1.0}} & - 
        & \multicolumn{1}{r|}{\textbf{1.0}} & \multicolumn{1}{r|}{\textbf{1.0}} & \multicolumn{1}{r|}{-} & \textbf{1.0} \\ \cline{2-14} 
\multicolumn{1}{l|}{} & \texttt{BanglaBERT} & 
        \multicolumn{1}{r|}{-} & \multicolumn{1}{r|}{\textbf{1.0}} & \multicolumn{1}{r|}{\textbf{1.0}} & -
        & \multicolumn{1}{r|}{\textbf{1.0}} & \multicolumn{1}{r|}{\textbf{1.0}} & \multicolumn{1}{r|}{\textbf{1.0}} & -
        & \multicolumn{1}{r|}{-} & \multicolumn{1}{r|}{-} & \multicolumn{1}{r|}{0.1} & - \\ \hline
\multicolumn{1}{l|}{\multirow{2}{*}{D18}} & \texttt{mBERT} & 
        \multicolumn{1}{r|}{\textbf{0.8}} & \multicolumn{1}{r|}{\textbf{1.0}} & \multicolumn{1}{r|}{\textbf{1.0}} & -
        & \multicolumn{1}{r|}{0.1} & \multicolumn{1}{r|}{\textbf{1.0}} & \multicolumn{1}{r|}{-} & \textbf{1.0} 
        & \multicolumn{1}{r|}{0.5} & \multicolumn{1}{r|}{\textbf{1.0}} & \multicolumn{1}{r|}{\textbf{1.0}} & - \\ \cline{2-14} 
\multicolumn{1}{l|}{} & \texttt{BanglaBERT} & 
        \multicolumn{1}{r|}{-} & \multicolumn{1}{r|}{\textbf{1.0}} & \multicolumn{1}{r|}{-} & \textbf{1.0}
        & \multicolumn{1}{r|}{-} & \multicolumn{1}{r|}{0.5} & \multicolumn{1}{r|}{0.5} & -
        & \multicolumn{1}{r|}{-} & \multicolumn{1}{r|}{\textbf{1.0}} & \multicolumn{1}{r|}{\textbf{1.0}} & - \\ \hline
\multicolumn{1}{l|}{\multirow{2}{*}{D19}} & \texttt{mBERT} & 
        \multicolumn{1}{r|}{-} & \multicolumn{1}{r|}{\textbf{1.0}} & \multicolumn{1}{r|}{-} & \textbf{1.0}
        & \multicolumn{1}{r|}{\textbf{1.0}} & \multicolumn{1}{r|}{\textbf{1.0}} & \multicolumn{1}{r|}{-} & \textbf{1.0} 
        & \multicolumn{1}{r|}{\textbf{1.0}} & \multicolumn{1}{r|}{\textbf{1.0}} & \multicolumn{1}{r|}{\textbf{1.0}} & - \\ \cline{2-14} 
\multicolumn{1}{l|}{} & \texttt{BanglaBERT} & 
        \multicolumn{1}{r|}{-} & \multicolumn{1}{r|}{\textbf{1.0}} & \multicolumn{1}{r|}{-} & \textbf{1.0}
        & \multicolumn{1}{r|}{-} & \multicolumn{1}{r|}{-} & \multicolumn{1}{r|}{-} & -
        & \multicolumn{1}{r|}{-} & \multicolumn{1}{r|}{\textbf{1.0}} & \multicolumn{1}{r|}{-} & \textbf{1.0} \\ \hline
\end{longtable}
\clearpage
\subsection{RQ2 Tables}
\begin{table}[!htb]
    \centering
    \caption{Fine-tuned BSA models' bias toward gender identity categories grouped by the BSA datasets' developers' gender identities.}
    \label{tab:dataset_developer_gender}
    \begin{tabular}{p{3cm}|p{2cm}|p{6cm}|p{1cm}}
    \hline
    \diagbox[width=3cm]{bias}{developer} &  \femaleemoji & \maleemoji & \femaleemoji+\maleemoji \\\hline
    \femaleemoji & 2 (D4m, D6m) & 4 \texttt{(D3m, D15B, D19m, D19B)} & 0 \\\hline
    \maleemoji  & 3 \texttt{(D6B, D11m, D11B)} & 12 \texttt{(D2m, D2B, D5m, D5B, D7m, D7B, D9m, D9B, D15m, D16m, D16B, D18m)} & 0 \\\hline
    no/rare & 1 (D4B) & 2 \texttt{(D3B, D18B)} & 0 \\\hline
    \end{tabular}
    
    \vspace{5pt}
    
    \centering
    \caption{Fine-tuned BSA models' bias toward religion-based identity categories grouped by the BSA datasets' developers' religious identities.}
    \label{tab:dataset_developer_religion}
    \begin{tabular}{p{3cm}|p{.75cm}|p{6.5cm}|p{1.9cm}}
    \hline
    \diagbox[width=3cm]{bias}{developer} &  \hinduemoji & \islamemoji & \islamemoji+Agnostic \\\hline
    \hinduemoji & 0 & 5 \texttt{(D4m, D15B, D16m, D18m, D19m)} & 1 \texttt{(D7m)} \\\hline
    \islamemoji & 0 & 13 \texttt{(D2m, D2B, D3B, D4B, D5m, D6B, D5B, D9m, D9B, D11m, D11B, D15m, D16B)} & 0 \\\hline
    no/rare & 0 & 4 \texttt{(D3m, D6m, D18B, D19B)} & 1 \texttt{(D7B)} \\\hline
    \end{tabular}
    
    \vspace{5pt}
    
    \centering
    \caption{Fine-tuned BSA models' bias toward nationality-based identity categories grouped by the BSA datasets' developers' national identities.}
    \label{tab:dataset_developer_nationality}
    \begin{tabular}{p{3cm}|p{8.5cm}|p{1.1cm}}
    \hline
    \diagbox[width=3cm]{bias}{developer} &  \bdflagemoji & \indiaflagemoji  \\\hline
    \bdflagemoji & 12 \texttt{(D2m, D3B, D4m, D5m, D7m, D7B, D9m, D9B, D11m, D16m, D16B, D19B)} & 0 \\\hline
    \indiaflagemoji  & 5 \texttt{(D2B, D6B, D18m, D18B, D19m)} & 0 \\\hline
    no/rare & 7 \texttt{(D3m, D4B, D5B, D6m, D11B, D15m, D15B)} & 0 \\\hline
    \end{tabular}
\end{table}
Each cell of these tables shows the number of fine-tuned BSA models that show bias toward identity category $x$ that developer(s) from identity category $y$ developed. Beside each count, we list the fine-tuned BSA models that fall into that criterion inside parentheses. To avoid repeating the base BERT models' names in the tables' cells, we used \texttt{Dxm} and \texttt{DxB}, respectively, to indicate the fine-tuned models resulting from training \texttt{mBERT} and \texttt{BanglaBERT} using the BSA dataset \texttt{Dx}.
\clearpage
\subsection{RQ3 Tables}
\begin{longtable}{p{0.5cm}|l|p{1cm}|p{1.75cm}|p{1cm}|p{1.75cm}|p{1cm}|p{1.75cm}}
\caption{Quantified Bias Metrics (average PCM and PCR) in ten data splits.}\label{tab:pcm_pcr}
\\\hline
\multicolumn{2}{l|}{Identity Dimension} & \multicolumn{2}{|c|}{\textbf{Gender}} & \multicolumn{2}{|c|}{\textbf{Religion}} & \multicolumn{2}{|c}{\textbf{Nationality}} \\\hline
\textbf{ID} & \textbf{Language Model} & PCM & PCR (\femaleemoji, \maleemoji) & PCM & PCR (\hinduemoji, \islamemoji) & PCM & PCR (\bdflagemoji, \indiaflagemoji) \\\hline
\endfirsthead
\caption*{\textbf{Table~\ref{tab:pcm_pcr} continued:} Quantified Bias Metrics (average PCM and PCR) in ten data splits.}
\\\hline
\multicolumn{2}{l|}{Identity Dimension} & \multicolumn{2}{|c|}{\textbf{Gender}} & \multicolumn{2}{|c|}{\textbf{Religion}} & \multicolumn{2}{|c}{\textbf{Nationality}} \\\hline
\textbf{ID} & \textbf{Language Model} & PCM & PCR (\femaleemoji, \maleemoji) & PCM & PCR (\hinduemoji, \islamemoji) & PCM & PCR (\bdflagemoji, \indiaflagemoji) \\\hline
\endhead
\multirow{2}{*}{D1} & \texttt{mBERT} &
    146.98 & 10, 0 & 104.7 & 0, 10 & 76.34 & 10, 0 \\ \cline{2-8}
    & \texttt{BanglaBERT} &
    79.97 & 0, 10 & 180.25 & 0, 10 & 62.61 & 10, 0 \\\hline
\multirow{2}{*}{D2} & \texttt{mBERT} &
    54.12 & 0, 10 & 31.57 & 0, 10 & 38.44 & 10, 0 \\ \cline{2-8}
    & \texttt{BanglaBERT} &
    71.82 & 0, 10 & 31.1 & 0, 10 & 37.66 & 1, 9 \\\hline
 \multirow{2}{*}{D3} & \texttt{mBERT} &
    55.46 & 10, 0 & 32.92 & 10, 0 & 45.89 & 1, 9 \\ \cline{2-8}
    & \texttt{BanglaBERT} &
    67.62 & 7, 3 & 33.23 & 7, 3 & 55.21 & 10, 0 \\\hline
\multirow{2}{*}{D4} & \texttt{mBERT} &
    92.04 & 10, 0 & 49.51 & 10, 0 & 50.44 & 10, 0 \\ \cline{2-8}
    & \texttt{BanglaBERT} &
    33.16 & 3, 7 & 11.14 & 1, 9 & 22.15 & 7, 3 \\\hline
 \multirow{2}{*}{D5} & \texttt{mBERT} &
    87.18 & 0, 10 & 47.46 & 0, 10 & 52.73 & 10, 0 \\ \cline{2-8}
    & \texttt{BanglaBERT} &
    58.48 & 0, 10 & 39.47 & 0, 10 & 24.9 & 4, 6 \\\hline
\multirow{2}{*}{D6} & \texttt{mBERT} &
    66.12 & 10, 0 & 24.69 & 7, 3 & 58.21 & 9, 1 \\ \cline{2-8}
    & \texttt{BanglaBERT} &
    110.49 & 0, 10 & 52.99 & 0, 10 & 81.21 & 0, 10 \\\hline
 \multirow{2}{*}{D7} & \texttt{mBERT} &
    76.23 & 0, 10 & 19.66 & 0, 10 & 46.34 & 10, 0 \\ \cline{2-8}
    & \texttt{BanglaBERT} &
    42.18 & 0, 10 & 22.43 & 0, 10 & 29.84 & 4, 6 \\\hline
\multirow{2}{*}{D8} & \texttt{mBERT} &
    42.35 & 0, 10 & 35.27 & 0, 10 & 46.51 & 10, 0 \\ \cline{2-8}
    & \texttt{BanglaBERT} &
    54.4 & 0, 10 & 29.04 & 0, 10 & 39.76 & 0, 10 \\\hline
 \multirow{2}{*}{D9} & \texttt{mBERT} &
    49.23 & 0, 10 & 64.55 & 0, 10 & 70.98 & 10, 0 \\ \cline{2-8}
    & \texttt{BanglaBERT} &
    75.62 & 0, 10 & 44.73 & 0, 10 & 31.36 & 10, 0 \\\hline
\multirow{2}{*}{D10} & \texttt{mBERT} &
    93.7 & 0, 10 & 62.63 & 0, 10 & 60.07 & 0, 10 \\ \cline{2-8}
    & \texttt{BanglaBERT} &
    48.51 & 0, 10 & 67.38 & 0, 10 & 67.21 & 0, 10 \\\hline
 \multirow{2}{*}{D11} & \texttt{mBERT} &
    7.28 & 0, 10 & 3.8 & 5, 5 & 6.26 & 10, 0 \\ \cline{2-8}
    & \texttt{BanglaBERT} &
    5.81 & 6, 4 & 2.62 & 2, 8 & 5.17 & 9, 1 \\\hline
\multirow{2}{*}{D12} & \texttt{mBERT} &
    26.52 & 3, 7 & 15.9 & 10, 1 & 25.9 & 1, 9 \\ \cline{2-8}
    & \texttt{BanglaBERT} &
    37.81 & 9, 1 & 14.41 & 9, 1 & 35.1 & 0, 10 \\\hline
 \multirow{2}{*}{D13} & \texttt{mBERT} &
    17.34 & 0, 10 & 9.94 & 2, 8 & 13.75 & 8, 2 \\ \cline{2-8}
    & \texttt{BanglaBERT} &
    4.46 & 10, 0 & 1.59 & 8, 2 & 7.05 & 0, 10 \\\hline
\multirow{2}{*}{D14} & \texttt{mBERT} &
    118.66 & 0, 10 & 26.31 & 10, 0 & 70.33 & 10, 0 \\ \cline{2-8}
    & \texttt{BanglaBERT} &
    108.36 & 0, 10 & 52.63 & 0, 10 & 50.43 & 10, 0 \\\hline
 \multirow{2}{*}{D15} & \texttt{mBERT} &
    58.25 & 0, 10 & 28.56 & 0, 10 & 46.22 & 2, 8 \\ \cline{2-8}
    & \texttt{BanglaBERT} &
    111.41 & 10, 0 & 38.55 & 10, 0 & 64.18 & 7, 3 \\\hline
\multirow{2}{*}{D16} & \texttt{mBERT} &
    29.79 & 0, 10 & 16.08 & 10, 0 & 67.04 & 10, 0 \\ \cline{2-8}
    & \texttt{BanglaBERT} &
    60.6 & 0, 10 & 20.86 & 0, 10 & 36.58 & 9, 1 \\\hline
 \multirow{2}{*}{D17} & \texttt{mBERT} &
    36.71 & 0, 10 & 90.19 & 0, 10 & 77.79 & 10, 0 \\ \cline{2-8}
    & \texttt{BanglaBERT} &
    96.24 & 0, 10 & 121.86 & 0, 10 & 48.57 & 2, 8 \\\hline
\multirow{2}{*}{D18} & \texttt{mBERT} &
    36.49 & 0, 10 & 10.19 & 10, 0 & 52.87 & 0, 10 \\ \cline{2-8}
    & \texttt{BanglaBERT} &
    59.48 & 9, 1 & 39.9 & 0, 10 & 32.27 & 0, 10 \\\hline
 \multirow{2}{*}{D19} & \texttt{mBERT} &
    39.28 & 3, 7 & 30.91 & 10, 0 & 51.45 & 0, 10 \\ \cline{2-8}
    & \texttt{BanglaBERT} &
    73.11 & 6, 4 & 30.6 & 0, 10 & 53.64 & 10, 0 \\\hline
\end{longtable}
\end{small}